\newcommand{\red}[1]{\textcolor{black}{#1}}

\documentclass[sigconf,twocolumn,10pt]{acmart}
\renewcommand\footnotetextcopyrightpermission[1]{}
\usepackage{graphicx} 

\usepackage{multicol,multirow}
\usepackage{array}
\usepackage{threeparttable}
\usepackage[ruled]{algorithm2e}
\usepackage{algpseudocode}
\usepackage{float}

\usepackage{booktabs}
\usepackage{makecell}

\usepackage{subcaption}
\usepackage{enumitem}
\usepackage{pifont}

\usepackage{tabularray}
\usepackage{amsmath}
\usepackage{booktabs} 

\usepackage{amssymb}

\makeatletter
\newcommand{\removelatexerror}{\let\@latex@error\@gobble}

\usepackage{balance}

\copyrightyear{2025}
\acmYear{2025} 
\setcopyright{cc}
\setcctype{by}

\def\name{\texttt{Cerberus}}
\def\namenormal{Cerberus}

\begin{document}
\title{Cerberus: Real-Time Video Anomaly Detection via Cascaded Vision-Language Models}
\author{
    Yue Zheng\textsuperscript{1},
    Xiufang Shi\textsuperscript{1},
    Jiming Chen\textsuperscript{2, 3}, 
    Yuanchao Shu\textsuperscript{2,\dag}}
    
\affiliation{%
  \institution{\textsuperscript{1}Zhejiang University of Technology, \textsuperscript{2}Zhejiang University, \textsuperscript{3}Hangzhou Dianzi University}
  \country{}
}


\begin{abstract}
Video anomaly detection (VAD) has rapidly advanced by recent development of Vision-Language Models (VLMs). While these models offer superior zero-shot detection capabilities, their immense computational cost and unstable visual grounding performance hinder real-time deployment. To overcome these challenges, we introduce \name{}, a two-stage cascaded system designed for efficient yet accurate real-time VAD. \name{} learns normal behavioral rules offline, and combines lightweight filtering with fine-grained VLM reasoning during online inference. The performance gains of \name{} come from two key innovations: motion mask prompting and rule-based deviation detection. The former directs the VLM's attention to regions relevant to motion, while the latter identifies anomalies as deviations from learned norms rather than enumerating possible anomalies. Extensive evaluations on four datasets show that \name{} on average achieves 57.68 fps on an NVIDIA L40S GPU, a 151.79$\times$ speedup, and 97.2\% accuracy comparable to the state-of-the-art VLM-based VAD methods, establishing it as a practical solution for real-time video analytics.
\end{abstract}


\thanks{\dag~Corresponding author: Yuanchao Shu (ycshu@zju.edu.cn).}
\maketitle

\section{Introduction}
\label{sec: introduction}
\begin{figure}[t!]
    \centering
    \includegraphics[width=0.98\linewidth]{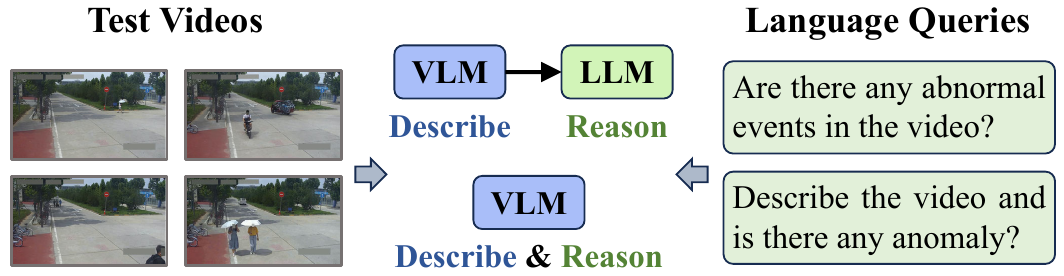}
    \vspace*{-2.4mm}
    \caption{Two common VLM-based VAD pipelines. Top: a modular two-step design where a VLM describes video content and an LLM reasons~\cite{zanella2024lavad,ye2025vera,yang2024anomlayruler}. Bottom: an integrated single-step design where a full-fledged VLM like Gemini~\cite{team2023gemini} and GPT-4o~\cite{achiam2023gpt4} handles both perception and reasoning~\cite{zhao2025smarthome,li2024videovista}.}
    \label{fig: VLM VAD example}
    \vspace*{-1em}
\end{figure}

Video anomaly detection (VAD) is a cornerstone task in video analytics that identifies unusual activities, such as traffic accidents or violent behaviors, with broad applications in public safety, traffic management, and smart surveillance~\cite{liu2025network_system_vad,ramachandra2020survey,sultani2018realworld}. The rise of large language models (LLMs) and vision-language models (VLMs) has opened new possibilities for VAD (Figure~\ref{fig: VLM VAD example}). 
Compared with conventional methods, VLM-based VAD offers two main advantages:

\begin{itemize}[leftmargin=0.4cm]
\item \textbf{From recognition to open-ended comprehension.}  
Traditional VAD relies on Deep Neural Networks (DNNs) that output data of predefined and fixed categories (e.g., object counts, bounding boxes, action labels). These systems can answer ``what is present'', but struggle to connect events into physical contexts. VLMs, by combining visual perception with linguistic knowledge from large-scale pretraining, enable deeper comprehension. They can perform causal inference, retrieve contextual details,  generate human-interpretable explanations, and hence provide a more flexible interface and finer granularity for anomaly detection. For example, instead of just detecting ``a running person'', a VLM can infer in what situation the person is running and whether it is abnormal.

\item \textbf{Flexible anomaly definition via natural language.}  
Traditional video analytics systems require complex and careful query planning, which requires extensive domain-specific experience~\cite{zhang2024vulcan,bhardwaj2022ekya,zhang2017videoStorm,jiang2018chameleon}. For example, detecting ``a person chasing another with a weapon'' may involve manual pipeline construction, tuning, and cross-platform deployment of modules including motion detector, object detectors, action recognizers, and trackers. On the other hand, VLM-based systems allow users to specify conditions directly in natural language, such as ``a person chasing another with a weapon in a crowded street''. It makes configuration simpler and more intuitive, leading to a lowered entry bar and bootstrapping cost.
\end{itemize}

While VLMs exhibit strong capabilities for VAD, their application presents critical challenges: 
\textbf{(1)~Prohibitive computational cost.} The massive scale of VLM architectures inherently demands substantial computational resources~\cite{zheng2024review}. In addition to the overhead incurred by the enormous size of the network, VLMs introduce extra overhead from cross-modal alignment~\cite{feng2025align,li2022mplug} and autoregressive decoding~\cite{you2024linear,bae2023fast}, both of which increase latency and memory usage. For example, in our experiments on an NVIDIA L40s GPU~\cite{nvidial40s}, processing 10 frames with Qwen2.5-VL-7B~\cite{wang2024qwen2vl} takes 8.48s and 17.85 GB of memory. This is about 20 times slower and heavier than modern DNN-based detectors like YOLOv10-L~\cite{wang2024yolov10}, which uses only 0.43s and 861 MB.
\textbf{(2)~Susceptibility to distraction in multimodal grounding.} Although natural language interfaces simplify anomaly specification, VLMs often show unstable grounding in complex scenes. This stems from pretraining, where alignment is shaped by captions that emphasize salient objects (e.g., large, bright, central), causing them to neglect subtle but critical cues~\cite{xiao2024can}. Additionally, spurious correlations and spatial-overlap heuristics can further divert attention to irrelevant regions~\cite{reich2024uncovering,zong2025ground}. In VAD, this is detrimental: for instance, a small peripheral car running a red light may be ignored if a nearby large bus dominates attention. 
\textbf{(3)~Lack of design in contextualizing VLMs for anomaly detection.} Conventional DNN-based pipelines excel at incorporating scene-specific priors, such as normality learned from background subtraction or trajectory clustering, which can enhance accuracy and adaptability across environments~\cite{jiang2021flexible,noghre2024exploratory}. However, the design of contextualizing VLM-based VAD methods is still in its infancy. Existing VLM-based VAD solutions rely mainly on models pretrained on general knowledge and prompt engineering~\cite{zanella2024lavad,ye2025vera}. Methods that attempt to improve accuracy by asking LLMs to enumerate possible anomalies~\cite{yang2024anomlayruler,ding2025slowfastvad} are also fragile. For example, in a traffic-monitoring scenario, a non-contextualized VLM which can identify ``accidents'' or ``assaults'' could easily overlook a contextual anomaly like ``skateboarding on a pedestrian-only lane''. 

To deal with these fundamental challenges and democratize VLMs for anomaly detection, we propose \name{}, a real-time VLM-based VAD system that combines lightweight perception and deep comprehension. The design of \name{} builds on three core mechanisms tailored for VLM and VAD: cascaded architecture, motion-mask prompting, and rule-based deviation detection. 

\textbf{First, to address prohibitive computational costs, we investigate how to minimize redundant inference.} Video streams are highly redundant and applying expensive multimodal alignment and decoding to all segments indiscriminately is extremely inefficient. Inspired by how humans skim ordinary scenes and only focus on unusual ones, we design \textit{a cascaded pipeline} that filters out redundant frames while keeps key semantic information. Specifically, a lightweight Contrastive Language-Image Pretraining (CLIP)~\cite{radford2021learning_CLIP} model performs coarse filtering to discard irrelevant frames, while a powerful VLM conducts fine-grained reasoning on the remaining candidates. The cascaded pipeline is carefully designed to allow the lightweight stage to trade precision for high recall, and hence anomalous content is preserved while the overhead is significantly reduced.

\textbf{Second, to mitigate distraction in multimodal grounding, we examine how to strengthen focus on relevant regions.} VLMs sometimes over-attend to salient but irrelevant elements, missing subtle cues that are decisive for VAD. It presents a critical question: how can we guide the model to focus on relevant regions? We observed that anomalies in videos are predominantly driven by foreground subject motion, while static regions contribute little anomaly signals. To this end, we propose \textit{motion mask prompting}, which uses temporal motion masks to highlight foreground activities and reduce background distractions in complex scenes.

\red{\textbf{Third, to achieve better contextualization in VAD, we investigate how to incorporate scene-specific knowledge effectively.}
Current VLM-based methods that either rely solely on pretrained knowledge or attempt to enumerate anomalies remain immature and often miss context-dependent events. This raises a key question: how can VLMs be infused with scene-specific context to achieve more reliable anomaly detection? Inspired by how scientific theories are distilled from repeated observations and then used to explain new phenomena, we consolidate ``what is normal'' from routine frames to induce contextual rules. Anomalies are then revealed as deviations from these rules. To this end, we introduce \textit{rule-based deviation detection}, which induces scene-specific norms offline and integrates them with VLM’s general knowledge during online inference.}


\name{} operates in two phases. In the offline phase, scene-specific normality rules are induced by combining VLM reasoning with LLM abstraction, with optional user customization. In the online phase, \textit{motion mask prompting} highlights foreground activities, which are then processed by a cascaded architecture: a CLIP-based model performs coarse-grained filtering, while a powerful VLM handles fine-grained reasoning. Within this cascaded design, \textit{rule-based deviation detection} is integrated into both filtering and reasoning to assess whether candidate events break offline-defined norms. 

\red{\name{} is implemented with state-of-the-art models including Qwen2.5-VL-7B and DeepSeek-R1-0528~\cite{guo2025deepseek_r1} for offline rule induction. During online inference, it employs PE-Core-L14-336 CLIP~\cite{bolya2025perception_encoder} for coarse-grained filtering and combines Qwen2.5-VL-7B with a Qwen3-Embedding-4B~\cite{zhang2025qwen3embedding} classifier for fine-grained reasoning\footnote{Note that the design of \name\ also works with other modern VLMs.}. Under a realistic setting where anomalies account for 1\% of frames on the most challenging NWPU Campus dataset~\cite{cao2023NWPUCampus}, the system achieves 45.81 FPS with a 138.8$\times$ end-to-end speedup while maintaining 97.2\% accuracy on par with the strongest baseline. In summary, our main contributions are as follows:}

\begin{itemize}[leftmargin=0.4cm]
    \item We propose \name{}, a real-time VAD system that enables natural language-based anomaly specification through integrated visual-language understanding.
    \item We design a cascaded architecture that combines lightweight CLIP-based filtering with VLM-based reasoning, greatly improving efficiency without sacrificing accuracy.
    \item We introduce \textit{motion mask prompting} to enhance grounding on motion-relevant regions and \textit{rule-based deviation detection} to capture anomalies as deviations of norms.
    \item We implement and evaluate \name{} on an edge server testbed, averaging 57.68 fps, a 151.79$\times$ speedup, and 97.2\% accuracy comparable to the best baseline.
\end{itemize}

\section{Background and Motivation}
\subsection{Vision-Language Models}
\begin{figure}[t!]
  \centering
    \subfloat[CLIP]
    {\includegraphics[width=0.304\linewidth]{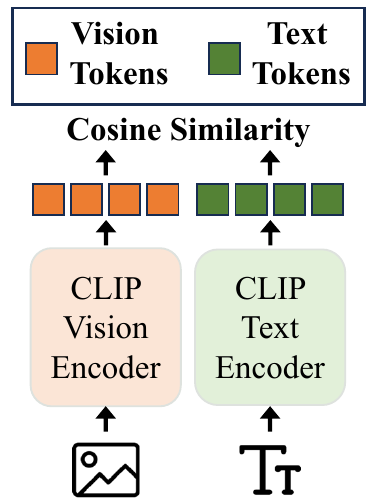}\label{subfig: background/VLM CLIP architecture a}}
    \hspace{1mm}
  \subfloat[Typical VLM]{
    \includegraphics[width=0.40\linewidth]{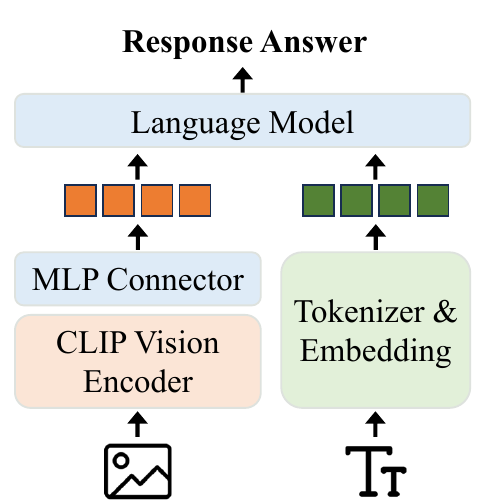}\label{subfig: background/VLM CLIP architecture b}}
      \vspace*{-3mm}
  \caption{Architectures of CLIP and a typical VLM.}
  \label{fig: background/VLM CLIP architecture}
  \vspace*{-1em}
\end{figure}

VLMs represent a major step in multimodal artificial intelligence, combining visual perception with natural language understanding through large-scale pretraining on image-text pairs~\cite{xiao2024can,reich2024uncovering,yan2025empowering}. Before the emergence of VLMs, CLIP established the foundation of vision-language alignment by using a dual-encoder design that independently encodes images and texts into a shared embedding space for contrastive learning~\cite{radford2021learning_CLIP} (Figure~\ref{subfig: background/VLM CLIP architecture a}). Building upon this, VLMs incorporate CLIP-style vision encoders with connector modules that align visual features to the text embedding space, and then leverage LLMs to generate responses~\cite{wang2024qwen2vl,zanella2024lavad} (Figure~\ref{subfig: background/VLM CLIP architecture b}). While both CLIP and VLMs process natural language and visual inputs, they serve different purposes. CLIP focuses on measuring image-text similarity without generating text, whereas VLMs excel at producing descriptive and reasoning-based textual outputs about visual content. To understand their potential and limitations for VAD applications, we conduct three sets of experiments that examine their representative properties.

\begin{table}[t!]
\centering
\resizebox{0.6\linewidth}{!}{
\begin{tabular}{ccc}
\toprule[2pt]
\textbf{Method} & \textbf{Type} & \textbf{AUC (\%)} \\
\midrule
GODS & I3D-RGB & 61.56  \\
RareAnom & I3D-RGB & 68.33 \\
PE-Core-L14-336 & CLIP & 64.31 \\
Qwen2.5-VL-7B & \textbf{VLM} & \textbf{82.51} \\
\bottomrule[2pt]
\end{tabular}
}
\vspace*{2mm}
\caption{Comparison of detection accuracy on a subset of \texttt{XD-Violence} dataset across different methods}
\label{tab: background/detection accuracy}
\vspace*{-6mm}
\end{table}

\noindent \textbf{Detection accuracy.}  
Detection accuracy in VAD is commonly evaluated using the Area Under the receiver operating characteristic Curve (AUC). As shown in Table~\ref{tab: background/detection accuracy}, conventional supervised methods like GODS~\cite{wang2019gods} and RareAnom~\cite{thakare2023rareanom}, which rely on I3D-RGB features~\cite{carreira2017quo_i3d}, achieve limited accuracy. While foundational models like PE-Core-L14-336 CLIP show comparable performance in a zero-shot setting, suggesting that large-scale vision-language pretraining provides a viable foundation for VAD, more advanced VLMs demonstrate a significant leap. For instance,  Qwen2.5-VL-7B achieves an 82.51\% AUC, substantially outperforming prior approaches and underscoring the strong potential of VLMs to generalize to unseen anomalous behaviors.

\noindent \textbf{Computational overhead.}  
As shown in Table~\ref{tab: background/computational overhead}, traditional models such as YOLOv10-L and Kinetics-I3D~\cite{carreira2017quo_i3d} process 10 frames in under 0.5s with less than 1.2 GB memory. CLIP requires about twice the time and three times the memory of YOLOv10-L. In contrast, Qwen2.5-VL-7B demands nearly 20$\times$ more time (8.48s vs.~0.43s) and memory (17.9~GB vs.~0.86~GB), making real-time deployment impractical. These results indicate that integrating the Transformer~\cite{vaswani2017attention} for joint vision–language reasoning introduces substantial overhead, whereas CLIP maintains a comparatively lightweight design compared to VLMs.

\begin{table}[t!]
\centering
\resizebox{0.7\linewidth}{!}{
\begin{tabular}{ccc}
\toprule[2pt]
\textbf{Method} & \textbf{Time (s)} & \textbf{Memory (GB)} \\
\midrule
YOLOv10-L & 0.43 & 0.86 \\
Kinetics-I3D & 0.38 & 1.18 \\
PE-Core-L14-336 & 0.84 & 3.19 \\
Qwen2.5-VL-7B & \textbf{8.48} & \textbf{17.85} \\
\bottomrule[2pt]
\end{tabular}
}
\vspace*{0.5em}
\caption{Computational overhead comparison on a NVIDIA L40S GPU for processing 10 frames.}
\label{tab: background/computational overhead}
\vspace*{-1.4em}
\end{table}

\begin{figure}[t!]
  \centering
    \subfloat[Input Frame]
    {\includegraphics[width=0.4\linewidth]{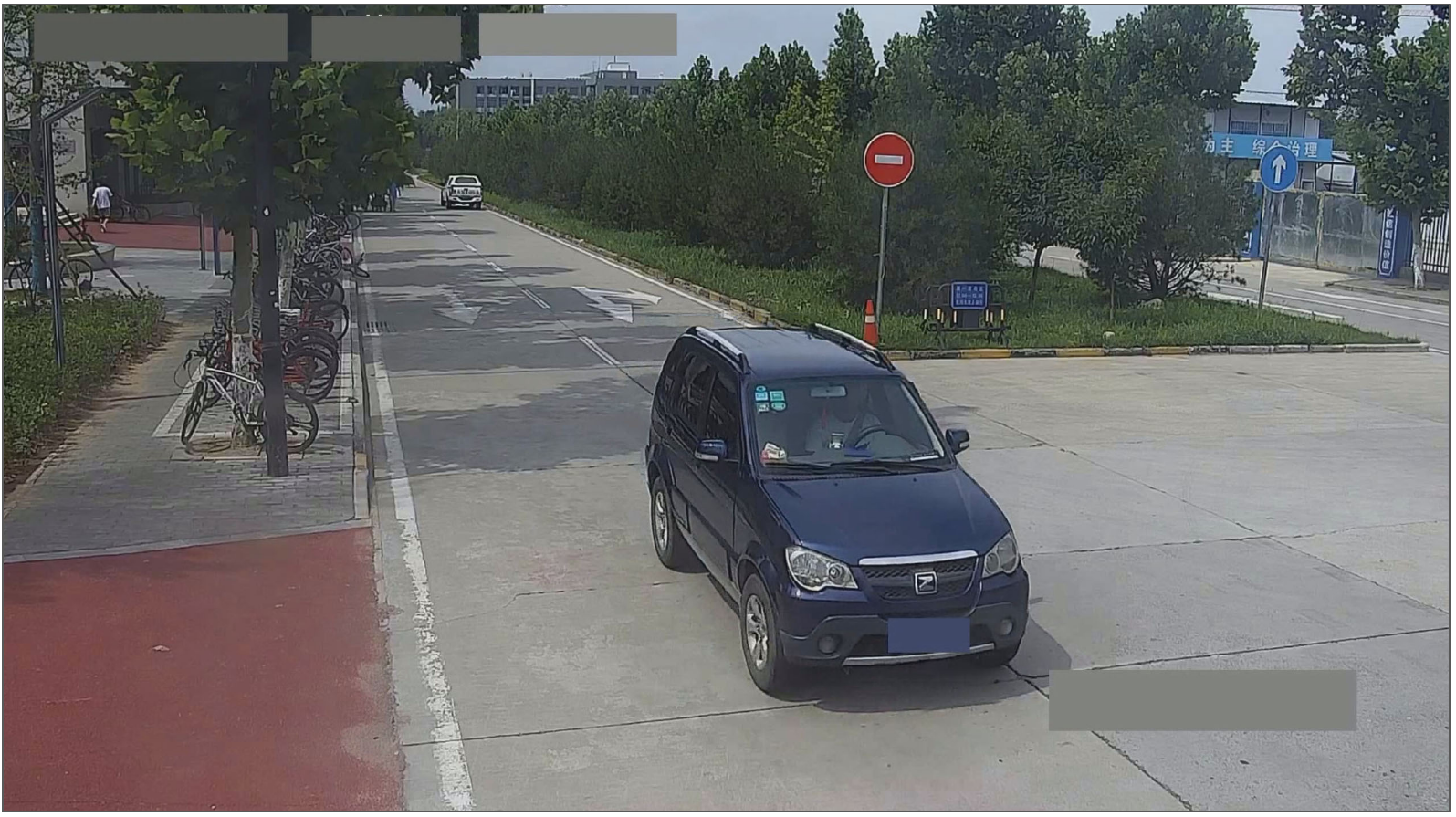}\label{subfig: background/distraction frame}\vspace{-1mm}}
    \hspace{4mm}
  \subfloat[VLM Attention Map]{
    \includegraphics[width=0.4\linewidth]{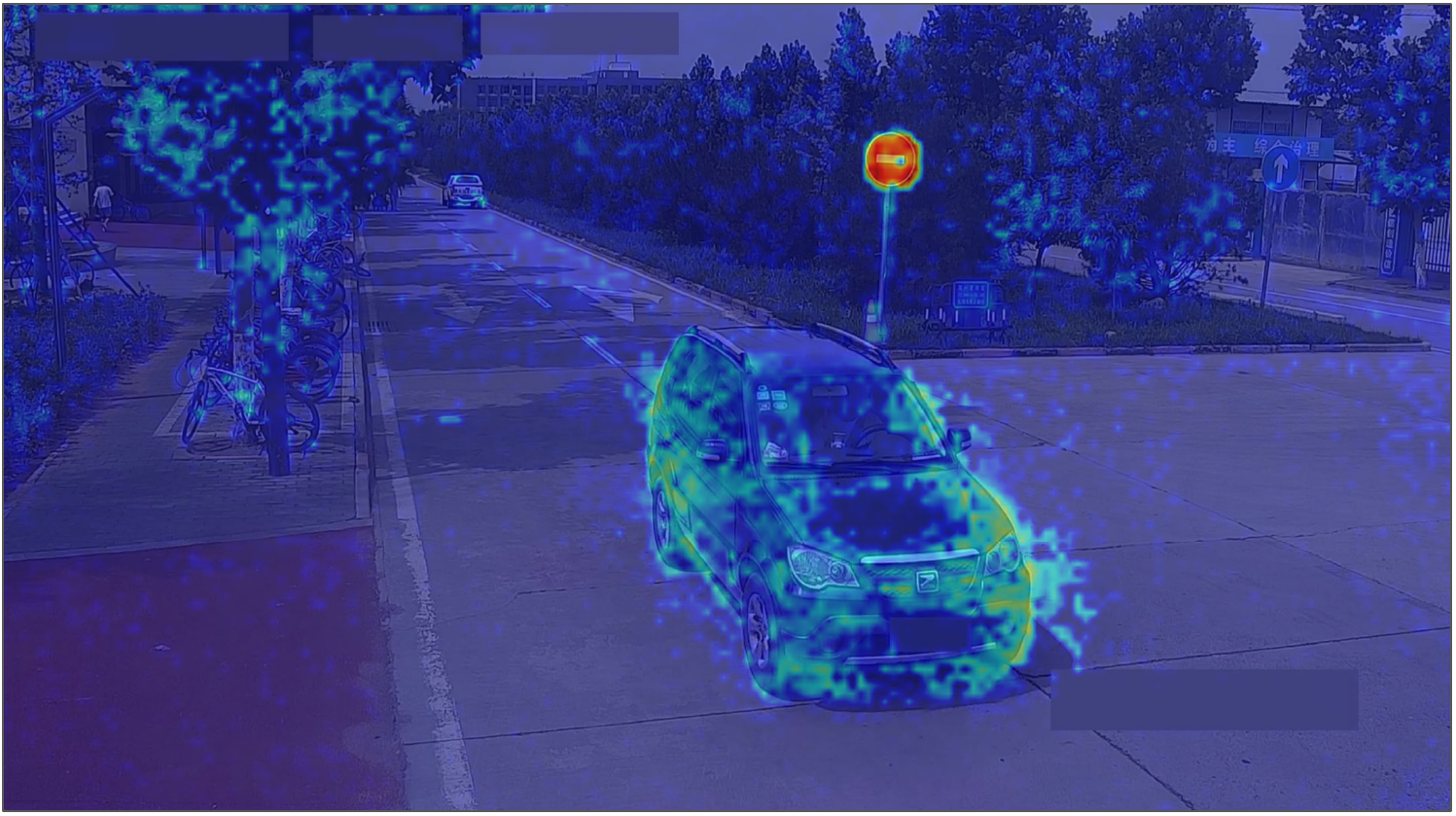}\label{subfig: background/distraction heatmap}\vspace{-1mm}}
          \vspace*{-2.6mm}
  \caption{An example of attentional distraction: the VLM focuses on the salient foreground objects, thereby missing crucial contextual cues like the traffic sign and the distant violating vehicle.}
  \label{fig: background/distraction}
    \vspace*{-3mm}
\end{figure}

\noindent \textbf{Susceptibility to distraction.} 
VLMs remain prone to attentional distraction due to immature cross-modal alignment and unstable attention mechanisms~\cite{xiao2024can,reich2024uncovering}. Figure~\ref{fig: background/distraction} illustrates a typical failure: when asked to determine whether a vehicle violates traffic rules based on the stop sign, Qwen2.5-VL-7B concentrated on the salient car and the sign, while its attention fragmented across irrelevant background details. Critically, it overlooked the white car driving in the wrong direction in the distance, the key element for answering the query. Such attention failures prevent accurate reasoning for VAD applications.

\subsection{Opportunities in Anomaly Detection}
\label{sec: background/real-world videos}
We ground our motivation in three key opportunities of real-world VAD tasks.

\begin{figure}[t!]
   \centering
   \includegraphics[width=.75\linewidth]{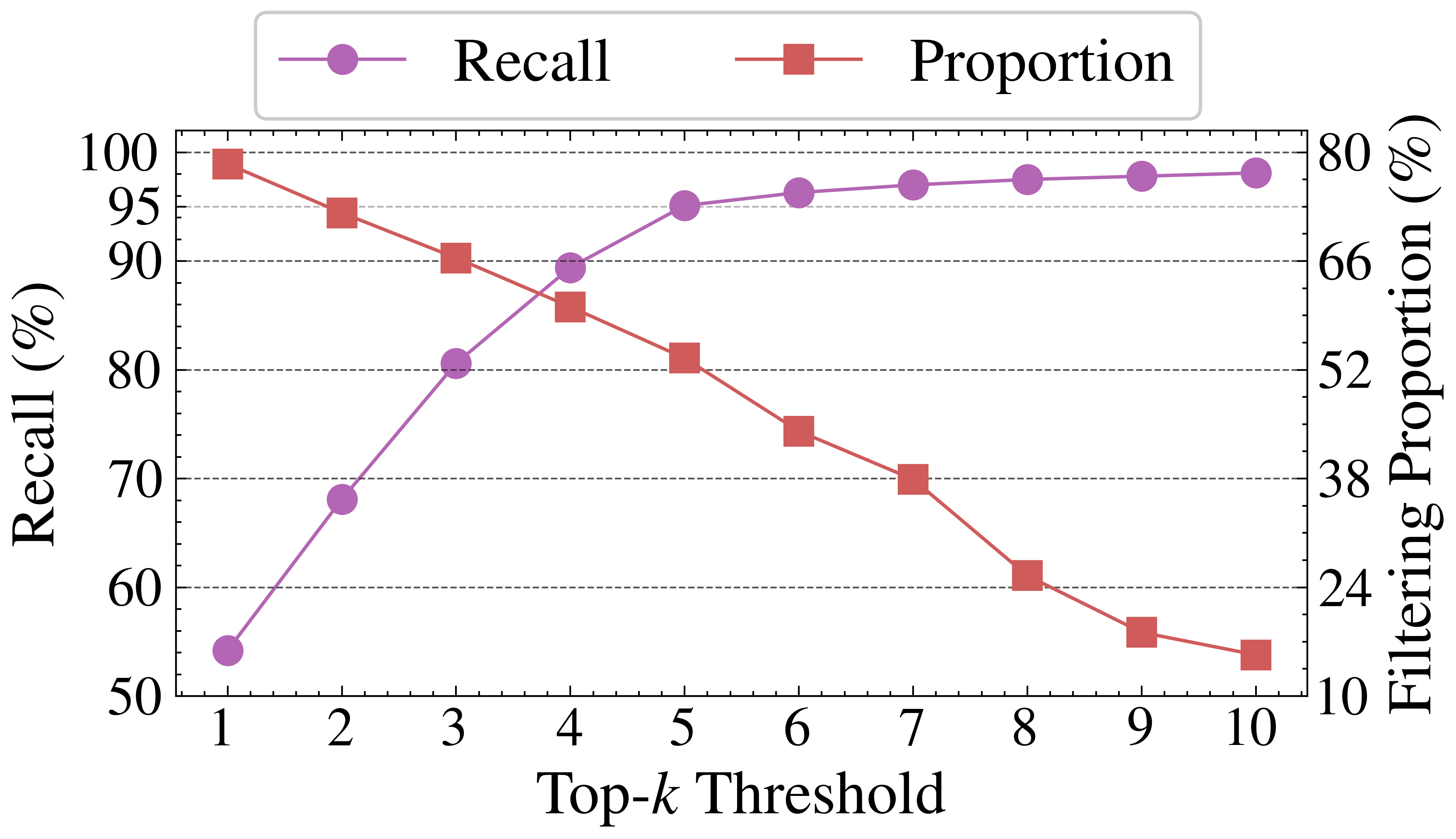}
   \vspace*{-4mm}
   \caption{Trade-off between anomaly recall and frame filtering proportion under different Top-$k$.}
   \label{fig: background/recall & proportion}
   \vspace*{-2mm}
\end{figure}

\noindent \textbf{Events Enable Cascaded Inference.}  
Real-world VAD datasets are dominated by normal events. For instance, anomalous frames account for only 5.38\% and 4.45\% in the ShanghaiTech (\texttt{SHTech})~\cite{luo2017SHTech} and NWPU Campus (\texttt{Campus})~\cite{cao2023NWPUCampus} datasets. This extreme imbalance implies that applying deep reasoning uniformly to all frames is wasteful, since most inputs are irrelevant to anomaly detection. To explore lightweight filtering, we applied PE-Core-L14-336 CLIP on the \texttt{SHTech} dataset. As shown in Figure~\ref{fig: background/recall & proportion}, setting the top-$k$ threshold to 5 preserves more than 95\% anomaly recall while discarding over 50\% of frames. This validates the feasibility of a front-end filter and motivates the cascaded inference strategy in our system.

\noindent \textbf{Foreground Motion Provides Reliable Cues.}  
Anomalies almost always involve motion, while static backgrounds contribute little useful information. Our temporal difference experiments show that removing static frames retains nearly all anomalies, yielding 99.91\% and 99.84\% recall on \texttt{SHTech} and \texttt{Campus}, respectively. This confirms motion as a dependable cue for anomaly localization. Moreover, prior studies on visual prompting demonstrate that highlighting salient objects (e.g., with red markers) improves VLM grounding ability~\cite{shtedritski2023does_red_circle}. Inspired by this, we use foreground motion as a natural guide to direct attention toward behaviorally relevant regions, thereby enabling more reliable reasoning.

\begin{table}[t!]
\centering
\resizebox{0.70\linewidth}{!}{
\begin{tabular}{cccc}
\toprule[2pt]
\textbf{Dataset} &\textbf{Precision (\%)} & \textbf{Recall (\%)} & \textbf{AUC (\%)} \\
\midrule
SHTech & 91.18 & \textbf{27.13} & 77.93 \\
Campus & 68.82 & \textbf{21.81} & 69.23  \\
\bottomrule[2pt]
\end{tabular}
}
\vspace{1mm}
\caption{Detection performance of anomaly-matching with anomalies enumerated by Deepseek-R1~\cite{guo2025deepseek_r1}.}
\label{tab: background/anomaly recall}
\vspace*{-8mm}
\end{table}

\noindent \textbf{Unbounded Anomalies Motivate Rule-based Detection.}  
Because VAD is inherently context-dependent, reliable detection requires environment-specific priors. A common approach is anomaly matching, where observed events are compared against a predefined anomaly set~\cite{yang2024anomlayruler}. While such priors provide partial knowledge, their coverage is fundamentally limited. As shown in Table~\ref{tab: background/anomaly recall}, anomaly-matching achieves reasonable AUC but suffers from low recall (below 30\%), leaving many anomalies undetected. This exposes the unreliability of enumeration-based strategies and highlights the urgent need for a more comprehensive approach: integrating scene-specific policies with VLMs’ universal knowledge. It motivates us to learn robust rules of normal behavior and detect deviations, enabling generalization to unseen anomalies without relying on fragile anomaly lists.

\section{System Overview}
\begin{figure*}[t!]
   \centering
   \includegraphics[width=.8\linewidth]{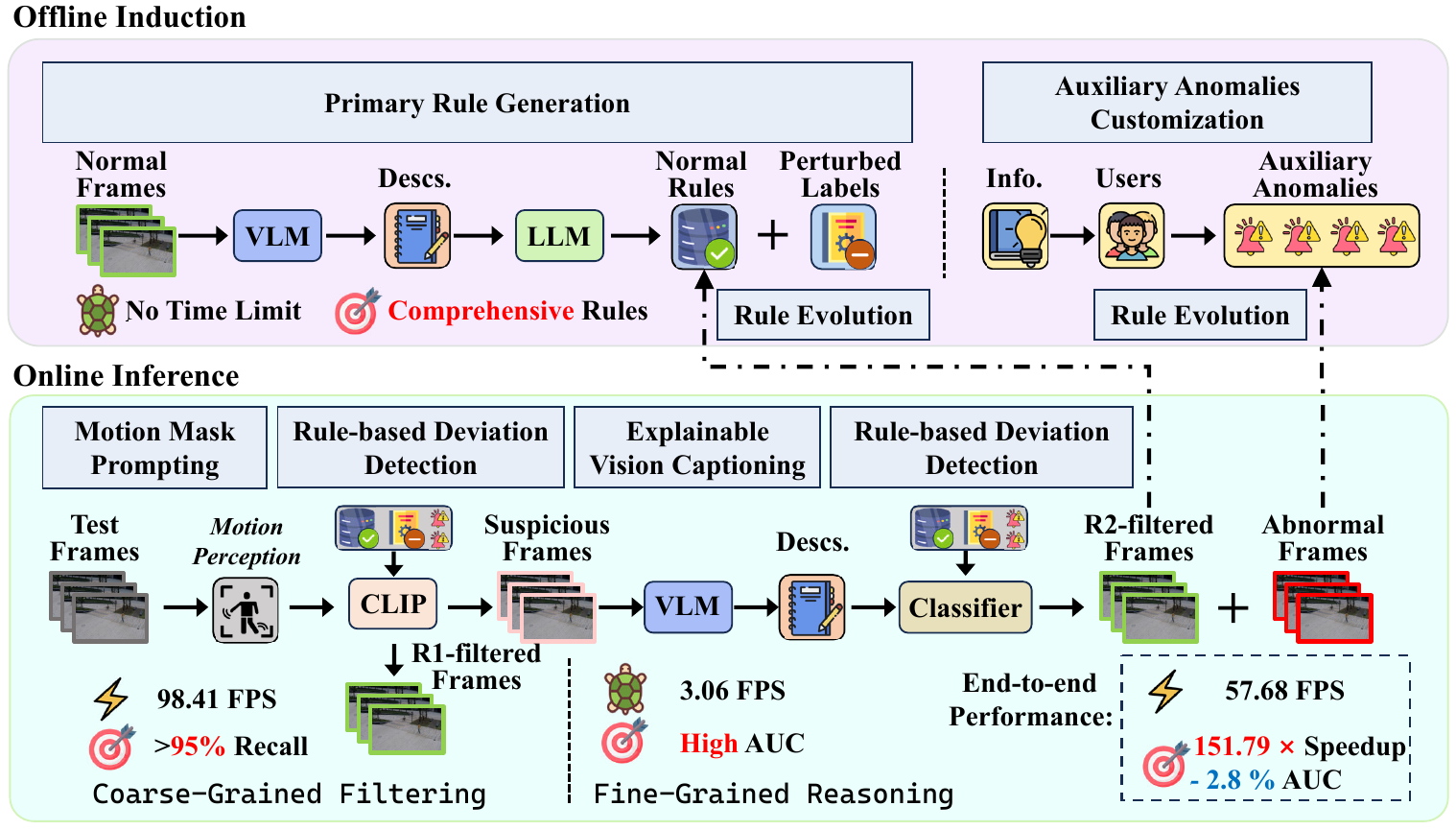}
   \captionsetup{width=.78\linewidth} 
   \caption{The system overview of \name{}. }
   \label{fig: system overview}
\end{figure*}

\name{} is an efficient, high-accuracy system for real-time VAD. As shown in Figure~\ref{fig: system overview}, it operates in two primary phases: an \textit{offline induction} phase to learn behavioral rules from sample videos, and an \textit{online inference} phase that uses these rules to efficiently detect anomalies in new video streams.

\noindent \textbf{Optimization Goal.} 
The final objective of \name{} is to minimize inference latency while maintaining detection performance comparable to that of a monolithic, fine-grained baseline. A straightforward baseline applies the accurate but slow fine-grained model $M_\text{F}$ to the entire video, achieving high accuracy at the cost of high latency. In contrast, \name{} introduces a fast, coarse-grained filter $M_\text{C}$ that operates at a much higher speed ($\bar{T}_{M_\text{C}} \ll \bar{T}_{M_\text{F}}$). This filter preemptively identifies and discards normal frames, passing only a small fraction, $\rho \in (0, 1]$, of suspicious frames to $M_\text{F}$ for detailed analysis. This system's performance is therefore optimized by maximizing the inference throughput subject to two critical constraints:
\vspace{-0.6em} 
\begin{align*}
    \max_{\rho} \quad & \text{Throughput}_{\text{\namenormal{}}} = \frac{1}{\bar{T}_{M_\text{C}} + \rho \cdot \bar{T}_{M_\text{F}}} \\
    \text{s.t.} \quad & \text{Recall}(M_\text{C}) \geq \theta, \\
                      & \text{AUC}(\text{\namenormal{}}) \geq \text{AUC}(\text{Baseline}) - \epsilon.
\end{align*}
\vspace{-1.6em} 

The first constraint ensures the coarse-grained filtering achieves very high recall ($\theta > 0.95$), minimizing the risk of discarding true anomalies. The second constraint guarantees that the overall AUC performance is nearly equivalent to the baseline, allowing for only a marginal tolerance $\epsilon$. This cascaded pipeline enables \name{} to achieve an average 151.79$\times$ speedup with only a 2.8\% decrease in AUC when the anomaly proportion is 1\%.

\noindent \textbf{Architecture.} The system's architecture is detailed below according to its two operational phases.

The \textit{offline induction} phase constructs a comprehensive rule base. It begins with \textit{primary rule generation} (\S\ref{subsec: primary rule generation}), where we leverage Qwen2.5-VL-7B to extract semantic descriptions of normal video segments, which are then abstracted into general behavioral rules by DeepSeek-R1-0528. To complement the positive normal rules, we introduce a pool of action labels that serve as perturbed references for anomaly detection. These labels are drawn from an external large-scale action dataset, ensuring broad semantic coverage without the need to enumerate anomalies explicitly.  Crucially, the \textit{auxiliary anomalies customization} (\S\ref{subsec: auxiliary anomalies customization}) module empowers supervisors to inject critical domain knowledge by manually defining specific violations. This entire rule base is kept current through a \textit{rule evolution} mechanism (\S\ref{subsec: rule evolution}), which integrates both automated and user feedback for refinement.

The \textit{online inference} phase employs an efficient two-tier cascade for real-time detection. First, the \textit{motion mask prompting} (\S\ref{subsec: motion masked prompting}) module intelligently filters out static frames and highlights dynamic regions of interest. These regions undergo a first-stage \textit{rule-based deviation detection} (\S\ref{subsec: rule-based deviation detection}) using PE-Core-L14-336 CLIP, which rapidly dismisses normal segments. Suspicious frames are then escalated to a fine-grained analysis stage. Here, an \textit{fine-grained captioning and detection} (\S\ref{subsec: explainable vision captioning and detection}) module with Qwen2.5-VL-7B generates interpretable scene descriptions. A second-stage rule-based deviation detection, powered by a Qwen3-Embedding-4B classifier, performs the final, precise anomaly confirmation. Confirmed anomalies feed back into the \textit{rule evolution} (\S\ref{subsec: rule evolution}) module, creating a dynamic learning loop that allows the system to adapt and improve over time.

\section{Offline Induction}
In this section, we introduce \textit{primary rule generation} that extracts behavioral norms from normal videos as scene-specific priors, \textit{auxiliary anomalies customization} for better user experience, and \textit{rule evolution} that refines the rule set through feedback mechanisms.
\subsection{Primary Rule Generation} \label{subsec: primary rule generation}
Despite recent progress, VLM-based approaches to VAD remain limited. Current methods often depend solely on VLMs' general knowledge without capturing scene-specific context, or rely on enumerating anomalies, which cannot be exhaustive. To achieve this transformation, \name{} employs a three-stage pipeline: (1) extracting behavioral descriptions from normal video segments, (2) abstracting these into generalizable rules, and (3) constructing a candidate pool that integrates normal rules with perturbed action labels for comprehensive VAD.

In the first stage, we extract normal video segments $S_{\text{normal}} = \{s_{\text{normal}_0}, ..., s_{\text{normal}_n}\}$, where each segment $s_{\text{normal}_i}$ comprises $k$ consecutive frames ($\{{f_{i, 1}, f_{i, 2}, ..., f_{i, k}}\}$) to preserve essential temporal dynamics. Using Qwen2.5-VL-7B, we process these segments with the structured prompt $p_\text{desc}$: \textit{``How many moving subjects (e.g., people, animals, vehicles) are in the scene, and what is each one doing in this specific scenario?''} This prompt is designed to elicit behaviorally meaningful descriptions that capture subject-environment relationships, moving beyond mere object detection. This process yields a corresponding textual description for each segment:
\begingroup
\setlength{\abovedisplayskip}{6pt} 
\setlength{\belowdisplayskip}{6pt} 
\begin{equation}
D_\text{normal} = \{\text{VLM}(s_{\text{normal}_i},p_{\text{desc}})|s_{\text{normal}_i}\in S_\text{normal}\}
\end{equation}
\endgroup

While these segment-level descriptions capture specific behavioral instances, they remain too granular for establishing scene-wide behavioral norms. To bridge this semantic gap, \name{} employs DeepSeek-R1-0528 to abstract these specific observations into a set of rules. This process is guided by the prompt $p_\text{rule}$: \textit{``Based on the following list of observed activities, summarize the general rules that define normal behavior in this scene. Focus on consistent actions, interactions, and locations.''} This operates through contextual inference (linking environmental cues with behavioral patterns) and pattern generalization (consolidating recurring observations). It can be expressed as follows:
\begingroup
\setlength{\abovedisplayskip}{6pt} 
\setlength{\belowdisplayskip}{6pt} 
\begin{equation}
    R_{\text{normal}}=\{\text{LLM}(D_{\text{normal}},p_\text{rule})\}
\end{equation}
\endgroup

Having established normal behavioral rules, we face a core challenge in VAD: the open-ended nature of anomalies makes exhaustive enumeration impossible, as any fixed set would remain incomplete and fail to capture novel events. A common workaround is to define a finite set of normal rules, $R_{\text{normal}}$, and then use exclusion-based rule matching. But this exclusion method often breaks down in complex scenes where normal and abnormal patterns coexist. For instance, a scene may display pedestrians walking on sidewalks (normal) while someone lies unconscious on the road (abnormal). Such cases reveal that conformity to normal rules does not guarantee the absence of anomalies.

To address both the infeasibility of enumerating anomalies and the limitations of relying solely on normal rules in mixed scenes, \name{} shifts the focus from rule enumeration and exclusion to detecting semantic deviations from established norms. The key insight is that anomalies typically diverge from normal patterns while aligning with diverse action concepts. To realize this, we augment the positive rule set with 339 atomic action labels  ($L_\text{perturbed}$) from the Moments in Time (\texttt{Moments}) dataset. These labels are particularly suited for this role: they comprehensively cover human, animal, and object-centered activities, form a highly clustered semantic space at atomic granularity for distinctions, and provide ready-to-use labels, thereby avoiding endless anomaly enumeration. This design creates a unified candidate pool:
\begingroup
\setlength{\abovedisplayskip}{6pt}
\setlength{\belowdisplayskip}{6pt}
\begin{equation}
    P_{\text{candidate}} = P_\text{perturbed}\bigcup P_\text{normal}
\end{equation}
\endgroup
where:
\begin{center}
\setlength{\abovedisplayskip}{-12pt}
\setlength{\belowdisplayskip}{6pt}
    \begin{align*}
        &P_\text{perturbed} = \{``\text{The scene depicts } \{l\}." \mid l \in L_{\text{perturbed}}\}. \\
        &P_\text{normal} = \{``\text{The normal scene depicts }\{r\}." \mid r \in R_{\text{normal}}\}.
    \end{align*}
\end{center}

This candidate pool enables semantic competition where anomalous content naturally exhibits higher similarity to perturbed labels than to scene-specific normal rules. The resulting candidate pool forms the foundation for \textit{rule-based deviation detection} (\S \ref{subsec: rule-based deviation detection}).

\subsection{Auxiliary Anomalies Customization} \label{subsec: auxiliary anomalies customization}
The automatically induced rules in \name{} capture behavioral norms effectively from visual patterns and work well in most scenarios. However, certain domain-specific constraints cannot be inferred from visual data alone. For example, time-based restrictions such as ``cycling is only allowed during daytime and not allowed at night'', or context-dependent policies like ``walking in prohibited directions during specific hours''. While these cases are relatively rare, they represent practical real-world requirements that pure visual analysis cannot address. To address them, \name{} provides a customization module that allows supervisors to add natural language rules reflecting domain-specific knowledge. These user-defined rules augment the automatically learned visual patterns to cover edge cases and improve overall detection completeness. Unlike traditional DNN-based VAD systems that required administrators to master both computer vision techniques and scene-specific knowledge, \name{} only asks them to express constraints in plain language, thereby lowering the barrier to use.

\subsection{Rule Evolution} \label{subsec: rule evolution}
To ensure sustained performance and adaptability, a \textit{rule evolution} module continuously refines the rule set using two complementary feedback loops from online inference (\S \ref{sec: online inference}).

\noindent \textbf{Fine-to-Coarse (F2C) Feedback}:
The fine-grained reasoning stage is the main computational bottleneck. To reduce its cost, \name{} reuses frames that were first marked as suspicious but later confirmed as normal by VLM (R2-filtered set in Figure~\ref{fig: system overview}). These hard negatives expose the weaknesses of the coarse filter. By adding them back into rule generation, the system learns more precise normal rules, allowing the coarse stage to discard more normal frames (R1-filtered set in Figure~\ref{fig: system overview}) and lighten the load on fine-grained reasoning.

\noindent \textbf{User-in-the-Loop (UIL) Feedback:}  
Automated detection may sometimes struggle with ambiguous cases near decision boundaries. For these, \name{} presents abnormal frames to users for validation and rule abstraction. This step goes beyond simple confirmation: users can generalize new anomaly rules from specific examples. Since abnormal contents are rare, this process adds little burden for supervisors but provides valuable semantic knowledge, enabling \name{} to steadily expand its detection capability.

\section{Online Inference} \label{sec: online inference}
In this section, \name{} processes incoming video streams in real-time through a cascaded architecture. It operates via coarse-grained filtering using \textit{motion mask prompting} and \textit{rule-based deviation detection}, followed by \textit{fine-grained captioning and detection} for precise and interpretable reasoning.

\subsection{Motion Mask Prompting} \label{subsec: motion masked prompting}
Long untrimmed videos often contain lengthy segments with static backgrounds. Sending all frames to VLMs wastes computation and dilutes attention from informative regions. Lengthy inactive periods (empty roads, idle hallways) consume resources without providing meaningful events. When activity occurs, relevant subjects are typically in the foreground, while static background elements can distract the model. To address these challenges, \name{} introduces \textit{motion masked prompting}, which leverages temporal differencing to drop static frames while providing motion-aware guidance for key regions. The approach computes frame-wise differences and determines motion proportion as:
\begingroup
\setlength{\abovedisplayskip}{6pt} 
\setlength{\belowdisplayskip}{6pt} 
\begin{equation}
    p(D_t) = \frac{\sum_{i=1}^{W} \sum_{j=1}^{H}\lvert F_t(i,j) - F_{t-1}(i,j) \rvert}{W\times H},
\end{equation}
\endgroup
where $F_t(i,j)$ represents pixel intensity at location $(i,j)$ and $W\times H$ denotes the total pixel count.

\begin{figure}[t!]
    \centering
    
    \subfloat[Original Frame]{\includegraphics[width=0.44\linewidth]{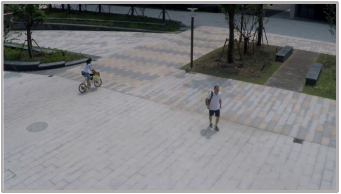}\label{fig: motion masked visual prompt a}\vspace{-2mm}}
    \hspace{4mm} 
    \subfloat[Motion Mask]{\includegraphics[width=0.44\linewidth]{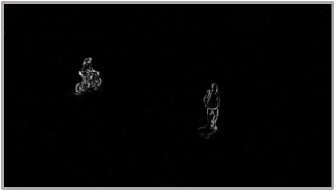}\label{fig: motion masked visual prompt b}\vspace{-2mm}}
    
    \vspace{1mm} 
    
    \subfloat[Red Circle Prompt]{\includegraphics[width=0.44\linewidth]{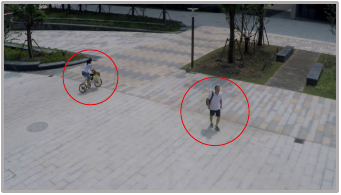}\label{fig: motion masked visual prompt c}\vspace{-2mm}}
    \hspace{4mm} 
    \subfloat[Red Square Prompt]{\includegraphics[width=0.44\linewidth]{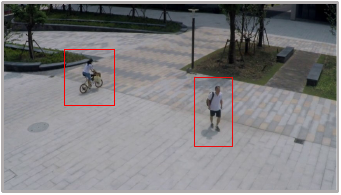}\label{fig: motion masked visual prompt d}\vspace{-2mm}}
    \vspace{-2mm}
    \caption{The generation process of \textit{motion mask prompting}. A motion mask (b) is derived from the original frame (a) to highlight moving subjects, which are then overlaid with red circles (c) or squares(d) prompts.}
    \label{fig: motion masked visual prompt}
    \vspace{-1em}
\end{figure}

Frames with $p(D_t)$ below the motion threshold $\epsilon$ are discarded as static. For the remaining frames, these values generate motion masks that localize regions by identifying pixels with significant changes. As illustrated in Figure~\ref{fig: motion masked visual prompt}, these masks serve as visual prompts by overlaying simple bounding cues (red circles or squares) on original frames. A single computation simultaneously handles both filtering and prompting, making the additional cost minimal.

Furthermore, recent work shows that simple visual cues have varying effectiveness in guiding CLIP and VLM attention: red circles demonstrate stronger attentional attraction, while red squares show moderate but still meaningful effects~\cite{shtedritski2023does_red_circle}. Building on this insight, \name{} adopts an adaptive prompting strategy that selects the cue type according to motion scale. A prompt-switching threshold $a$ separates subtle from prominent motions:  
\textbf{Red circles} ($\epsilon < p(D_t) < a$) highlight small or distant movements that could otherwise be overlooked. By emphasizing subtle activity, they improve the model's attention to fine details. 
\textbf{Red squares} ($p(D_t) \ge a$) capture large or elongated subjects more effectively. Rectangular bounding avoids including excessive background, thereby improving filtering efficiency and reducing distraction in downstream reasoning.  

Together, these complementary cues balance sensitivity and efficiency: circles ensure high recall by capturing subtle signals, while squares improve precision by suppressing background noise. 

\subsection{Rule-based Deviation Detection} \label{subsec: rule-based deviation detection}
After \textit{motion mask prompting}, candidate frames must be evaluated against contextual norms. Instead of enumerating anomalies, \name{} evaluates each segment against both scene-specific rules and perturbed labels in the candidate pool $P_{\text{candidate}}$ established in \textit{primary rule generation}.

The process, detailed in Algorithm~\ref{alg: clip-rule-deviation}, begins by encoding the visual features of the segment $s$ and each text description $t \in P_{\text{candidate}}$ into a shared embedding space using a CLIP-based model. To focus on the most informative evidence, we select the top-$k$ candidates $C_{\text{top-}k}(s)$, ranked by their cosine similarity scores. A health score $S(s)$ is calculated by aggregating these scores:

\begin{figure}[t!]
   \centering
   \includegraphics[width=.98\linewidth]{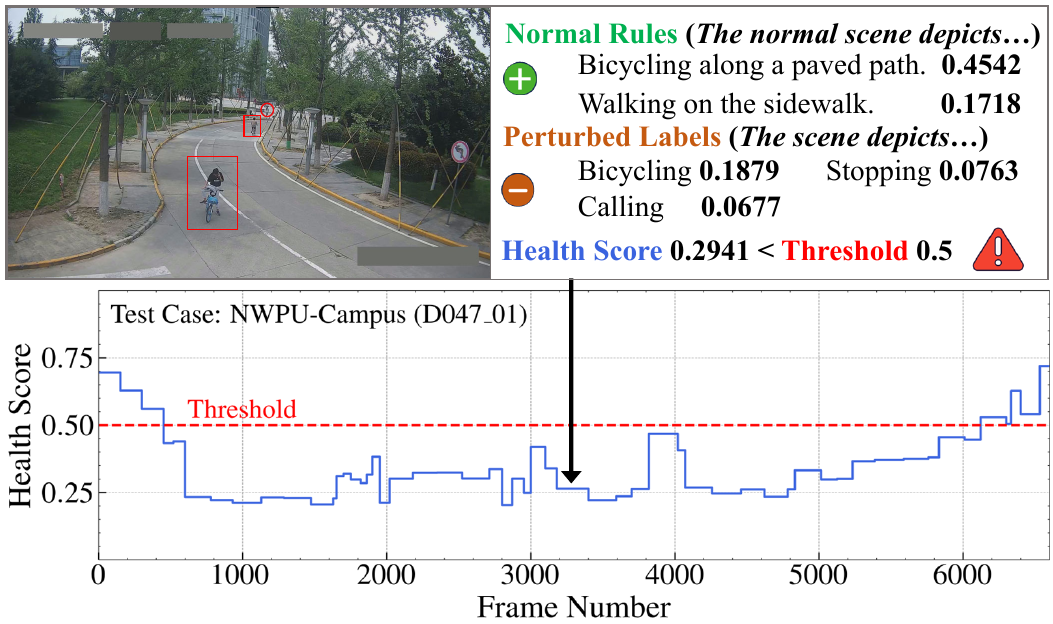}
   \vspace{-2mm}
   \caption{Example of rule-based deviation detection for VAD. The score is the difference between normal rules and perturbed labels, compared against the threshold.}
   \label{fig: rule-based deviation detection}
\end{figure}

\begingroup
\setlength{\abovedisplayskip}{10pt}   
\setlength{\belowdisplayskip}{6pt} 
\begin{equation}
S(s) = \sum_{t \in C_{\text{top-}k}(s)} w_t \cdot \text{sim}(v_s, v_t),
\label{eq: health score}
\end{equation}
\endgroup
where the weight $w_t = +1$ if $t$ is a normal rules ($t \in P_\text{normal}$) and $w_c = -1$ if it is a perturbed label ($t \in P_\text{perturbed}$). This formulation effectively rewards alignment with normal behavior while penalizing correspondence with perturbed labels. A segment is classified as anomalous if its health score $S(s)$ falls below a predefined threshold $\tau$.

Figure~\ref{fig: rule-based deviation detection} illustrates this mechanism in action. The health score remains above the threshold during normal events, as similarity to scene-specific normal rules dominates. When an anomaly occurs, the score drops sharply because the observed content deviates significantly from predefined scene-specific norms, naturally exhibiting higher semantic similarity to perturbed labels within the action space. This design offers a significant advantage: it detects anomalies by exploiting the fact that deviating behaviors naturally become more similar to perturbed descriptions in semantic space, rather than matching explicitly predefined anomalous patterns. This allows the system to identify unforeseen anomalies through semantic competition between normal and perturbed descriptions. In our implementation, we set $k=5$.

\begingroup
\begin{algorithm}[t!]
\caption{Deviation-driven anomaly detection with CLIP}
\label{alg: clip-rule-deviation}
\LinesNumbered
\resizebox{0.9\linewidth}{!}{
\begin{minipage}{\linewidth} 
\KwIn{Segment $s$, Candidate pool $P_\text{candidate} = P_{\text{normal}} \cup P_{\text{perturbed}}$, $\text{CLIP model}$, threshold $\tau$, top-$k$;}
\KwOut{Detection results (\textit{normal} / \textit{abnormal});}

$v_s \gets \text{CLIP.encode\_image}(s)$;

\For{$t \in P_\text{candidate}$}{
    $v_t \gets \text{CLIP.encode\_text}(t)$;
    
    $\text{sim}(t) \gets \cos\big(v_s, v_t\big)$;
}

Select top-$k$ candidates $C_{\text{top-}k}(s)$ ranked by $\text{sim}(\cdot)$;

\For{$t\in C_{\textnormal{top-}k}(s)$}{
    $w_t \gets +1$ \textbf{if} $t\in P_{\text{normal}}$ \textbf{else} $-1$;
}

$S(s) \gets \sum_{t\in C_{\text{top-}k}(s)} w_t \cdot p_t$; 

\eIf{$S(s) < \tau$}{
    \Return \textit{abnormal};
}{
    \Return \textit{normal};
}
\end{minipage}
}
\end{algorithm}
\endgroup

\subsection{Fine-grained Captioning and Detection} \label{subsec: explainable vision captioning and detection}
While the coarse-grained filtering achieves a high recall (over 95\%) for suspicious frames, it sacrifices precision by retaining many normal frames. Relying solely on this set for final decisions would generate excessive false alarms.

To address these limitations, \name{} employs a fine-grained reasoning stage that combines \textit{explainable vision captioning} with \textit{rule-based deviation detection}. Unlike the previous stage that directly processes visual content with CLIP, this stage first leverages a VLM to generate comprehensive textual descriptions of suspicious frames. These captions capture visual elements, actions, and contextual cues, providing an interpretable intermediate layer.

The textual representations are then evaluated using the same health scoring mechanism from Equation~\ref{eq: health score}, with a key distinction: similarity is computed entirely in text space. A text embedding model (e.g., Qwen3-Embedding) measures semantic alignment between VLM-generated captions and the candidate pool $P_{\text{candidate}}$, generating the final detection.

This decoupled architecture provides three advantages: 
\textbf{(1)~Specialization:} the VLM brings strong visual understanding capabilities, generating comprehensive scene descriptions, while the text embedding model specializes in semantic similarity and rule-based scoring, allowing each component to operate at its best capacity. \textbf{(2)~Interpretability:} converting visual evidence into explicit language makes decisions transparent rather than black-box; 
\textbf{(3)~Modularity:} each component can be independently upgraded without system-wide modifications, ensuring long-term adaptability.

The overall cascaded architecture balances recall and precision: coarse-grained filtering retains all potential anomalies while removing redundant frames, and fine-grained analysis provides interpretable reasoning for suspicious events, ensuring efficient and reliable online inference.

\section{Evaluation}
This section evaluates \name{} against state-of-the-art baselines on detection accuracy and overhead, and conducts ablation studies to assess each module's contribution.

\subsection{Experimental Setup}
\noindent \textbf{Implementation.}
\name{} runs on an edge server (Intel Xeon Platinum 8352V, 64GB RAM, NVIDIA L40S GPU) with Ubuntu 20.04, PyTorch 2.6.0, and CUDA 12.4. The offline phase uses Qwen2.5-VL-7B for visual captioning and DeepSeek-R1-0528 for rule generalization. Online detection employs OpenCV~\cite{opencv} temporal differencing for motion masks, PE-Core-L14-336 CLIP for coarse filtering, Qwen2.5-VL-7B for fine-grained captioning, and Qwen3-Embedding-4B for final classification.

\begin{table}[t!]
\centering
\resizebox{0.7\linewidth}{!}{
\begin{tabular}{lccc}
\toprule[2pt]
\makecell{\textbf{Dataset}} & 
\makecell{\textbf{\# Testing} \\ \textbf{Frames}} & 
\makecell{\textbf{\# Anomaly} \\ \textbf{Classes}} & 
\makecell{\textbf{Anomaly} \\ \textbf{Ratio}} \\  
\midrule
Avenue & 15,324 & 5 & 25.23\% \\ 
SHTech & 42,883 & 11 & 42.47\% \\ 
UBnormal & 92,640 & 22 & 74.53\% \\ 
Campus & 384,059 & 28 & 16.63\% \\
\bottomrule[2pt]
\end{tabular}
}
\vspace{0.5em}
\caption{Description of the VAD test datasets used.}
\label{tab: datasets description}
\vspace{-2em}
\end{table}

\noindent \textbf{Baselines.} 
We compare \name{} to the following alternatives: 
(1) \texttt{AnomalyRuler}~\cite{yang2024anomlayruler}: Employs a VLM to describe frames and an LLM to verify them with offline-induced rules.
(2) \texttt{AnomalyRuler-base}~\cite{yang2024anomlayruler}: A variant of AnomalyRuler that replaces LLM verification with keyword matching.
(3) \texttt{CLIP with Rules}: A CLIP model utilizing AnomalyRuler's offline rules.
(4) \texttt{VLM with Rules}: utilizing AnomalyRuler's offline rules; 
Fair Comparison, all baselines are tested with the same auxiliary anomalies. The CLIP and VLM in the baselines use the same configuration as in \name{}.

\begin{table*}[t!]
    \setlength{\abovecaptionskip}{0pt}
    \setlength{\belowcaptionskip}{6pt}
  \centering
  \begin{threeparttable}
  \resizebox{\linewidth}{!}{
\begin{tabular}{ccccccccccc}
    \toprule[2pt]
    \multirow{3}{*}{\textbf{\makecell{Methods}}} & \multicolumn{2}{c}{\textbf{Avenue}} & \multicolumn{2}{c}{\textbf{SHTech}} & \multicolumn{2}{c}{\textbf{UBnormal}} & \multicolumn{2}{c}{\textbf{Campus}} & \multicolumn{2}{c}{\textbf{Avg.}} \\
    \cmidrule(lr){2-3} \cmidrule(lr){4-5} \cmidrule(lr){6-7} \cmidrule(lr){8-9} \cmidrule(lr){10-11}
    & \begin{tabular}[c]{@{}c@{}}Relative\\ AUC\end{tabular} & \begin{tabular}[c]{@{}c@{}}Throughput\\ (fps)\end{tabular} & \begin{tabular}[c]{@{}c@{}}Relative\\ AUC\end{tabular} & \begin{tabular}[c]{@{}c@{}}Throughput\\ (fps)\end{tabular} & \begin{tabular}[c]{@{}c@{}}Relative\\ AUC\end{tabular} & \begin{tabular}[c]{@{}c@{}}Throughput\\ (fps)\end{tabular} & \begin{tabular}[c]{@{}c@{}}Relative\\ AUC\end{tabular} & \begin{tabular}[c]{@{}c@{}}Throughput\\ (fps)\end{tabular} & \begin{tabular}[c]{@{}c@{}}Relative\\ AUC\end{tabular} & \begin{tabular}[c]{@{}c@{}}Throughput\\ (fps)\end{tabular} \\
    \midrule
    AnomlayRuler \textbf{(Mean)}\tnote{1}         & 100\%     & 0.46 & 100\%     & 0.41 & 100\%     & 0.34 & 100\%     & 0.33 & 100\%     & 0.38 \\
    AnomlayRuler-base \textbf{(Mean)}    & 91.63\%   & 0.82 & 92.31\%   & 0.76 & 93.07\%   & 0.63 & 88.43\%   & 0.66 & 91.36\%   & 0.72 \\
    VLM with Rules \textbf{(Mean)}      & 88.64\%   & 3.08 & 90.44\%   & 3.39 & 82.38\%   & 2.58 & 86.13\%   & 3.55 & 87.15\%   & 3.15\\
    CLIP with Rules \textbf{(Mean)}     & 71.32\%   & 118.32   & 77.45\%   & 112.61  & 73.29\%   & 87.82  & 68.84\%   & 90.06   & 72.73\%   & 102.29 \\
    \toprule
    \namenormal{} \textbf{(Orig.)}~\tnote{2}             & {96.87\%} & {4.76} & {98.13\%} & {4.90} & {96.84\%} & {3.02} & {97.13\%} & {6.21} & 97.24\% & 4.74 \\
    \namenormal{} \textbf{(5\%)}           & 97.15\% & 17.24 & 98.11\% & 32.49 & 97.23\% & 28.09 & 96.84\% & 17.79 & 97.33\% & 23.96 \\
    \namenormal{} \textbf{(1\%)}              & {96.82\%} & {53.97} & {97.66\%} & {72.25} & {97.15\%} & {57.54} & {97.21\%} & {45.81} & {97.21\%} & {57.68} \\
    \bottomrule[2pt]
    \end{tabular}
    }
\begin{tablenotes}
\item[1] \textbf{(Mean)}: Mean performance, since results showed minimal variation across different abnormal proportions.
\item[2] \textbf{(Orig.)}, \textbf{(5\%)}, or \textbf{(1\%)}: Denotes the abnormal frame proportions for \name{} in the test set.
\end{tablenotes}
\end{threeparttable}
\caption{Comparison of relative AUC and throughput for different VAD methods.}
\label{tab: performance_comparison}
\end{table*}

\begin{figure*}[htbp]
   \centering
   \includegraphics[width=.98\linewidth]{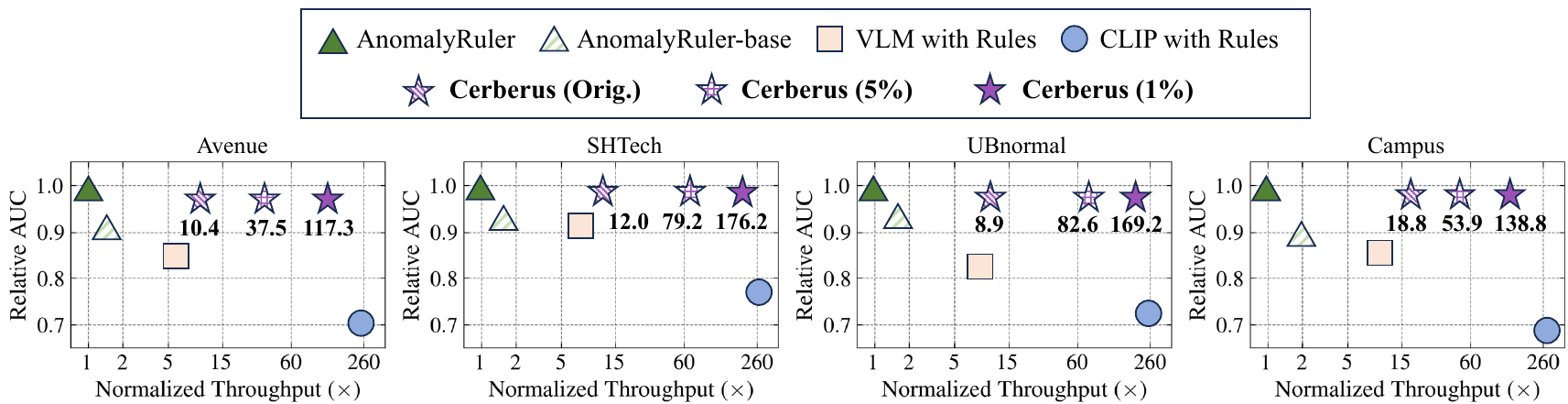}
   \vspace{-3mm}
   \caption{Illustration of normalized throughput vs. relative AUC on different datasets. }
  \label{fig: fps_vs_auc performance}
\end{figure*}

\noindent \textbf{Metrics.} 
We evaluate the performance of \name{} using the following metrics:
(1)~\textit{AUC}: AUC is the primary metric for detection accuracy, following standard practice in VAD tasks. 
(2)~\textit{Throughput} and \textit{Overhead}: Throughput measures the average frame rate (frames per second, fps), while overhead reflects the total time cost of a processing stage. 
(3)~\textit{Filtering Proportion}: It quantifies the proportion of frames filtered out during coarse-grained filtering. To ensure anomalies are not mistakenly removed, it is only meaningful when recall exceeds 95\%. 
(4)~\textit{Recall} and \textit{Precision}: Recall measures how well the system avoids missed detections (abnormal frames not recognized), while precision reflects how well it avoids false alarms (normal frames mistakenly flagged as abnormal). 
Notably, key detection metrics, including \textit{AUC}, \textit{recall}, and \textit{precision}, are measured in \textbf{percentages (\%)}, where higher values up to 100\% indicate better performance.

\noindent \textbf{Datasets.} \label{subsubsec: datasets} We evaluate \name{} on four semi-supervised VAD datasets: (1)~CUHK Avenue (\texttt{Avenue})~\cite{lu2013Avenue}, a single-scene dataset; (2)~\texttt{SHTech}, consisting of 13 campus scenes; (3)~\texttt{UBnormal}~\cite{acsintoae2022UBnormal}, a large-scale synthetic benchmark with 29 virtual scenes spanning streets, pavements, beaches, and airports; and (4)~\texttt{Campus}, the most challenging dataset featuring strong context-dependent anomalies. The dataset statistics in Table~\ref{tab: datasets description} show substantially higher anomaly proportions than those typically observed in real-world deployments, where anomalies are far less frequent. To better reflect deployment scenarios, we construct additional evaluation sets with reduced anomaly proportions using stratified normal frame duplication. For each dataset, we provide three configurations: \textbf{Original} (unaltered), \textbf{5\%}, and \textbf{1\%} anomaly proportions, with the original version used in all experiments unless otherwise specified. Notably, the weakly-supervised datasets such as UCF-Crime~\cite{sultani2018real_UCF_Crime} and XD-Violence~\cite{wu2020not_XD_Violence} are excluded, as their video-level labels do not align with our offline induction from normal segments.

\vspace{-1em}
\subsection{Overall Performance}
Table~\ref{tab: performance_comparison} provides a general comparison across four public benchmarks. Existing approaches face a key trade-off between accuracy and throughput. \texttt{AnomalyRuler} achieves perfect accuracy (100\% Relative AUC), but operates at impractical 0.46 fps. \texttt{CLIP with Rules} delivers high speed but suffers substantial accuracy loss. \texttt{AnomalyRuler-base} and \texttt{VLM with Rules} fall between these extremes. \name{} breaks this trade-off through two key innovations:

\noindent \textbf{Throughput Improvement:} \name{}'s coarse-grained filtering stage employs motion temporal difference and lightweight CLIP to identify and remove most normal frames early in the pipeline. This design achieves remarkable speed improvements by reserving expensive processing only for suspicious sets. The method improves throughput from 0.38 fps to 4.74 fps, representing a 12.47$\times$ acceleration over \texttt{AnomalyRuler}. Notably, under realistic conditions with low anomaly proportions (5\% and 1\%), the throughput further increases to 23.96 fps and 57.68 fps, respectively.

\noindent \textbf{Accuracy Enhancement:} \name{} leverages \textit{motion mask prompting} and \textit{rule-based deviation detection} to achieve superior detection accuracy. Compared to methods that similarly avoid LLM-based result double-check (\texttt{AnomalyRuler-base} and \texttt{VLM with Rules}), \name{} achieves 6\% and 10\% higher accuracy. When compared to \texttt{AnomalyRuler} with LLM-based verification, it narrows the accuracy gap to only 2.79\%. This strong performance stems from two factors: first, motion mask prompting focuses attention on foreground objects, reducing background distractions that could impair detection; second, it can systematically detect deviations from normal behavioral rules, whereas exhaustive anomaly enumeration can be incomplete and miss edge cases.

The throughput-accuracy trade-off is visualized in Figure~\ref{fig: fps_vs_auc performance}, where \name{} consistently occupies the optimal top-right region with both high accuracy and throughput. In contrast, \texttt{AnomalyRuler} achieves the highest accuracy but the lowest throughput, while \texttt{CLIP with Rules} offers the inverse. Under realistic low-anomaly conditions, \name{} demonstrates exceptional acceleration with at least 117.3$\times$ and 37.5$\times$ speedup at 5\% and 1\% anomaly proportions, respectively, by efficiently filtering abundant normal frames early in the pipeline.

\begin{table}[t!]
\centering
\resizebox{0.9\linewidth}{!}{
\begin{tabular}{lccccc}
\toprule[2pt]
\makecell{\textbf{Method}} & 
\makecell{\textbf{\# Rules}}&  
\makecell{\textbf{Precision}}& 
\makecell{\textbf{Recall}} & 
\makecell{\textbf{AUC}}\\
\midrule
\textbf{\namenormal{}} & \textbf{10.67} & \textbf{89.34} & \textbf{48.24} & \textbf{82.73} \\ 
w/o. Action Captioning & -2.34 & -35.71 & -11.23 & -18.56 \\ 
w/o. Context Captioning & -1.00 & -5.27 & -2.45 & -6.13 \\ 
w/o. Rule Generalization & +2.66 & -8.21 & -14.29 & -14.92 \\ 
\bottomrule[2pt]
\end{tabular}
}
\vspace{0.5em}
\caption{Ablation study of rule generation components: action captioning, context captioning, and rule generalization on \texttt{SHTech} dataset.}
\label{tab: ablation rule generation}
\vspace{-1em}
\end{table}

\subsection{Evaluation of Offline Induction}
We systematically evaluate each key component of \name{}, beginning with the offline induction stage.

\subsubsection{Effect of Rule Generation}
The \textit{primary rule generation} of \name{} integrates visual captioning and rule generalization. We conduct ablation experiments on the \texttt{SHTech} dataset by randomly selecting an equal number of normal frames under different configurations for rule generation and report the average over three runs. As shown in Table~\ref{tab: ablation rule generation}, captioning with action- and context-related information is critical: when either source is removed, the resulting rules become incomplete, leading to fewer valid rules. This incompleteness leads to significant precision degradation (up to 35.71\% drop), where normal behaviors are incorrectly flagged as anomalous due to the dominance of perturbed labels in health scoring. In contrast, rule generalization enriches the rule set and significantly boosts recall. Without it, rules remain overly specific and fail to capture broader behavioral patterns, causing true anomalies to be overlooked, leading to a 14.29\% recall drop. These complementary effects highlight that all three components: action cues, context cues, and generalization, are indispensable for robust VAD.

\subsubsection{Impact of Anomaly Customization}
The \textit{auxiliary anomaly customization} module addresses cases where constraints are not directly observable from visual cues. We configure domain-specific rules for two representative scenarios: the \texttt{Avenue} dataset, where people walking toward or away from the camera are considered anomalous despite no visible traffic violations, and the \texttt{SHTech} dataset, where prolonged loitering is labeled anomalous although it visually resembles harmless walking behavior. The corresponding customized rules are ``walking toward or away from the camera is anomalous'' and ``loitering is anomalous''. As shown in Table~\ref{tab:ablation_customization}, adding such rules consistently improves both AUC and throughput. For example, \texttt{Avenue} and \texttt{SHTech} achieve 2.28\% and 3.82\% AUC gains, while throughput also increases since domain-specific rules reduce false positives early in the pipeline. These results demonstrate that customization effectively complements automatically induced rules, extending coverage to special cases.

\begin{table}[t!]
\centering
\resizebox{1\linewidth}{!}{
\begin{tabular}{ccccc}
\toprule[2pt]
\multirow{3}{*}{\textbf{Method}} & \multicolumn{2}{c}{\textbf{Avenue}} & \multicolumn{2}{c}{\textbf{SHTech}} \\
\cmidrule(lr){2-3} \cmidrule(lr){4-5}
&AUC&\makecell{Throughput (fps)} & AUC & \makecell{Throughput (fps)} \\
\midrule
\makecell[l]{\namenormal{} (Base)} & {86.40} & {4.76} & {82.73} & {4.90} \\
\makecell[l]{\namenormal{} (\textbf{Customized})} & \textbf{+2.28} & \textbf{+0.22} & \textbf{+1.82} & \textbf{+0.31} \\
\bottomrule[2pt]
\end{tabular}
}
\vspace{1mm}
\caption{Impact of \textit{auxiliary anomaly customization}, comparing \name{} with and without customized anomalies on \texttt{Avenue} and \texttt{SHTech} datasets.}
\label{tab:ablation_customization}
\vspace{-2em}
\end{table}


\begin{figure}[t!]
\centering
\includegraphics[width=0.9\linewidth]{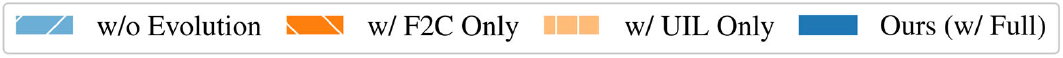}
  \subfloat[Effectiveness Analysis]{\includegraphics[width=0.46\linewidth]{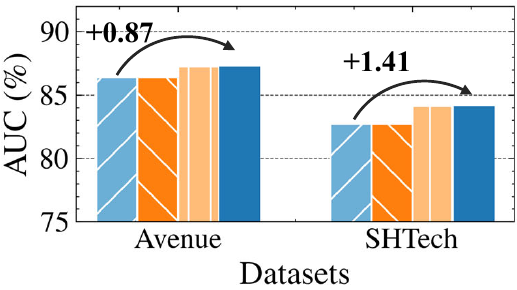}\label{subfig: ablation rule evolution a}\vspace{-1mm}}
      \hspace{1mm}
  \subfloat[Efficiency Analysis]{\includegraphics[width=0.46\linewidth]{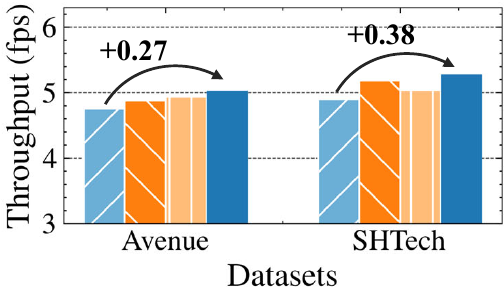}\label{subfig: ablation rule evolution b}\vspace{-1mm}}
  \vspace{-1mm}
    \caption{Impact of \textit{rule evolution} feedback mechanisms on AUC and throughput, showing individual and combined effects on \texttt{Avenue} and \texttt{SHTech} datasets.}
    \label{fig: ablation rule evolution}
\end{figure}

\subsubsection{Role of Rule Evolution}
\textit{Rule evolution} employs two complementary feedback mechanisms: F2C and UIL. Each dataset (\texttt{Avenue} and \texttt{SHTech}) is randomly split into two equal halves. The system first performs inference on one half, then applies rule evaluation and feedback updates, and finally tests the updated rules on the other half. In F2C feedback, normal frames identified by fine-grained reasoning are fed back to the \textit{primary rule generation} to refine rules. In UIL feedback, anomalies are added to the \textit{auxiliary anomalies customization} to expand the anomaly set. To avoid bias from prior knowledge of the dataset, Qwen-VL-Max~\cite{bai2023qwenvlversatilevisionlanguagemodel} is used to simulate user customization. Results are averaged over three runs with different random splits. As shown in Figure~\ref{fig: ablation rule evolution}, the two feedback mechanisms are complementary: F2C accelerates inference by improving throughput with little effect on accuracy, while UIL enhances accuracy by leveraging user feedback and also contributes to throughput gains. Combining both yields the best trade-off: on \texttt{Avenue}, accuracy improves by 0.87 and FPS by 0.27; on \texttt{SHTech}, accuracy improves by 1.41 and FPS by 0.38. These results confirm that both feedback mechanisms jointly enable \name{} to become progressively more accurate and efficient.

\subsection{Evaluation of Online Inference}
Next, we evaluate each module in the online inference stage.

\begin{table}[t!]
\centering
\resizebox{0.80\linewidth}{!}{
\begin{tabular}{lccc}
\toprule[2pt]
\makecell{\textbf{Method}} &
\makecell{\textbf{Anomaly Prop.}} &
\makecell{\textbf{AUC}}&
\makecell{\textbf{Overhead (h)}} \\
\midrule

\multirow{3}{*}{\makecell[l]{Coarse-grained\\filtering only}} & Orig. & 67.85 & 0.01 \\
& 5\% & 67.84 & 0.09 \\
& 1\% & {67.85} & {0.45} \\
\cmidrule{1-4}

\multirow{3}{*}{\makecell[l]{Fine-grained\\reasoning only}} & Orig.& 84.24 & 0.35 \\
& 5\% & 84.24 & 2.98 \\
& 1\% & {84.23} & {14.63} \\
\cmidrule{1-4}

\multirow{3}{*}{\makecell[l]{\textbf{\namenormal{}}\\\textbf{(Both stages)}}} & Orig. & {82.73} & {0.23} \\
& 5\% & 82.72 & 0.30 \\
& 1\% & {82.71} & {0.69} \\
\bottomrule[2pt]
\end{tabular}
}
\vspace{0.5em}
\caption{Ablation study of cascaded architecture components showing AUC performance and overhead comparison under different anomaly proportions.}
\label{tab: ablation-coarse-fine-stage}
\vspace{-1.2em}
\end{table}

\subsubsection{Impact of Cascaded Architecture}
We evaluate the contributions of coarse-grained filtering and fine-grained reasoning on a subset of the \texttt{SHTech} dataset, where 10\% of frames are sampled to form the original set. Anomalies are preserved, and two additional versions with anomaly proportion of \textbf{5\%} and \textbf{1\%} are constructed by duplicating normal frames (detailed in \S \ref{subsubsec: datasets}). As shown in Table~\ref{tab: ablation-coarse-fine-stage}, the fine-grained reasoner alone consistently achieves the highest accuracy (84.2\% AUC), but its overhead increases sharply as the dataset grows larger with more duplicated normal frames, making it impractical for deployment. Conversely, the coarse-grained filter remains extremely efficient, but its accuracy remains much lower (67.9\% AUC). Our \name{} provides a favorable balance: under the same anomaly proportions, it retains accuracy close to fine-grained reasoning while reducing overhead. Notably, in more realistic settings with fewer anomalies (5\% and 1\%), the benefit of \name{} becomes more pronounced. Its overhead stays lightweight while its accuracy far surpasses coarse-grained filtering, making it more suitable for real-world deployment.

\begin{figure}[t!]
\centering
\includegraphics[width=0.9\linewidth]{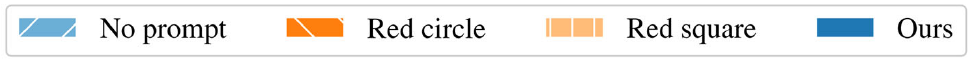}
  \subfloat[Coarse-grained filtering proportion]{\includegraphics[width=0.48\linewidth]{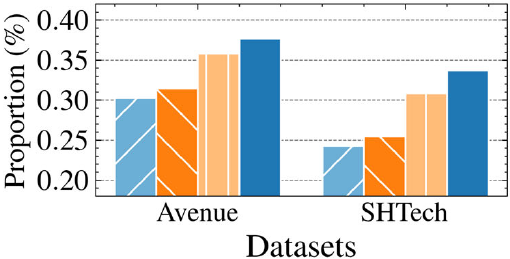}\label{subfig: ablation motion mask prompting a}\vspace{-2mm}}
      \hspace{1mm}
  \subfloat[Fine-grained reasoning accuracy]{\includegraphics[width=0.48\linewidth]{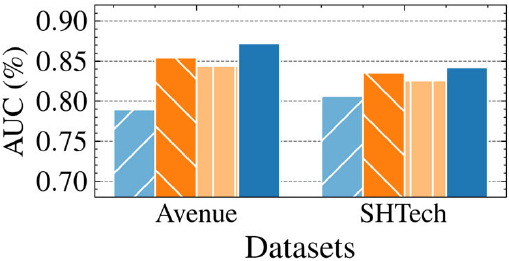}\label{subfig: ablation motion mask prompting b}\vspace{-2mm}}
  \vspace{-2mm}
  \caption{Comparison of different motion mask prompts on filtering efficiency and reasoning accuracy. In all cases, the recall of anomalies in coarse-grained filtering remains above 95\%.}
    \label{fig: ablation motion mask prompting}
\end{figure}

\begin{figure}[t!]
    \centering
 \subfloat[Effect of motion threshold on performance through motion sensitivity]{\includegraphics[width=0.48\linewidth]{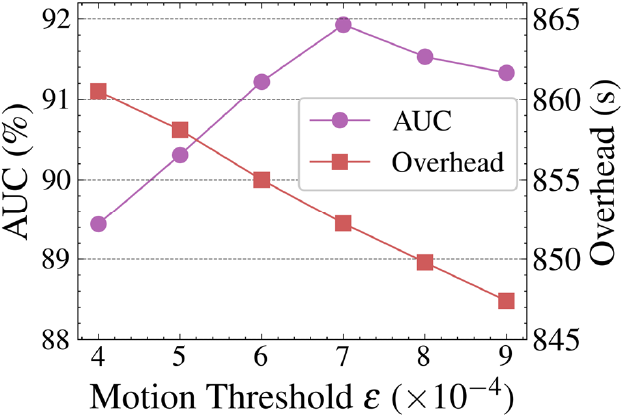}\label{subfig: ablation difference threshold a}\vspace{-1mm}}
      \hspace{2mm}
  \subfloat[The accuracy-efficiency trade-off with the prompt-switching threshold]{\includegraphics[width=0.48\linewidth]{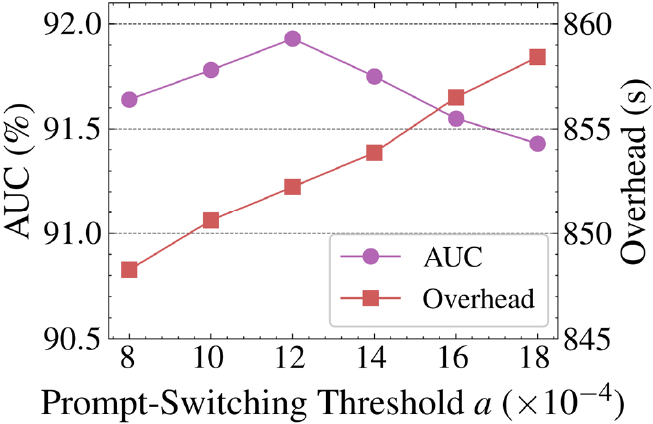}\label{subfig: ablation difference threshold b}\vspace{-1mm}}
  \vspace*{-1mm}
    \caption{The performance of \name{} with thresholds for \textit{motion mask prompting} on a subset of \texttt{SHTech}.}
    \label{fig: ablation difference threshold}
\end{figure}

\subsubsection{Trade-offs in Motion Mask Prompting} 
We first evaluate the \textit{motion mask prompting} module with different types of visual prompts. As shown in Figure~\ref{fig: ablation motion mask prompting}, the red circle prompt achieves higher fine-grained accuracy, while the red square prompt performs better in coarse-grained filtering, and our approach balances these complementary strengths. This result can be explained as follows. Red circles reduce missed detections by strongly highlighting subtle or distant motions, but they may introduce false positives that weaken coarse filtering. In contrast, red squares provide tighter spatial coverage that suppresses background distractions for coarse filtering, but they may overlook small or subtle actions, leading to lower fine-grained accuracy. To address this trade-off, our method applies red circles to subtle subjects to reduce missed detections and red squares to dominant subjects to suppress background distractions, achieving better overall performance.  

We further study two critical thresholds that influence performance. The motion detection threshold $\epsilon$ determines the minimum activation for motion: values too low may include noisy pixels with meaningless fluctuations, while values too high discard frames containing subtle but important motion, both degrading system accuracy. The prompt-switching threshold $\alpha$ controls the transition between prompt types and directly explains the observed trade-offs. When $\alpha$ is too high, the system tends toward using ``red circles'' predominantly, filtering fewer frames, and resulting in higher overhead while introducing more false detections that reduce AUC. When $\alpha$ is too low, the system trends toward using ``red squares'' predominantly, filtering more frames to reduce inference overhead but also overlooking smaller anomaly details, leading to decreased accuracy. Figure~\ref{fig: ablation difference threshold} presents the experimental results for these thresholds. Our analysis confirms that $\epsilon= 7\times10^{-4}$ and $\alpha=1.2\times10^{-3}$ provide an optimal trade-off, achieving 91.93\% AUC with 852.24s overhead by strategically combining the strengths of both prompt types while minimizing their respective weaknesses.
  
\subsubsection{Comparison of Rule-based Detection Methods}
We evaluate the \textit{rule-based deviation detection} module by comparing \name{} with \texttt{AnomalyRuler-base}. This baseline represents a typical anomaly-matching approach that enumerates possible anomalies using a reasoning LLM. As shown in Table~\ref{tab: background/anomaly recall}, our method achieves 21.11\% and 19.16\% higher recall on \texttt{SHTech} and \texttt{Campus}, respectively, resulting in AUC gains of 4.80\% and 4.52\%. Although precision decreases slightly (about 1\%), this is mainly due to the coarse-grained filtering step, which accelerates inference by discarding normal frames but may occasionally remove valid anomalies. Overall, \name{} substantially improves end-to-end detection performance by mitigating the anomaly omission problem inherent in existing methods.

\begin{table}[t!]
\centering
\resizebox{0.86\linewidth}{!}{
\begin{tabular}{lcccccc}
\toprule[2pt]
\multirow{2}{*}{\textbf{Method}} & \multicolumn{3}{c}{\textbf{SHTech}} & \multicolumn{3}{c}{\textbf{Campus}} \\
\cmidrule(lr){2-4} \cmidrule(lr){5-7}
& Precision & Recall & AUC & Precision & Recall & AUC \\
\midrule
\makecell[l]{Anomaly-\\matching}  & 91.18 & 27.13 & 77.93 & 68.82 & 21.81 & 69.23 \vspace{1mm}\\
\textbf{\namenormal{}{}} & {89.34} & \textbf{48.24} & \textbf{82.73} & {68.21}& \textbf{40.97}& \textbf{73.75}\\
\bottomrule[2pt]
\end{tabular}
}
\vspace{0.5em}
\caption{End-to-end detection performance of different rule-based detection methods.}
\label{tab: ablation rule-based detection methods}
\vspace{-1em}
\end{table}

\section{Related Work}

\noindent \textbf{Video Anomaly Detection.} 
Existing approaches are commonly categorized by supervision level. Supervised~\cite{liu2022traffic_fully_supervised,aboah2021vision_fully_supervised} and weakly supervised methods~\cite{li2022self_weakly_supervised,li2022scale_weakly_supervised,tian2021weakly_weakly_supervised} rely on detailed annotations, which are costly given that anomalies are rare and context-dependent. Unsupervised~\cite{wu2022self_unsupervised,wang2021robust_unsupervised,zaheer2022generative_unsupervised} and one-class approaches~\cite{liu2021hybrid_one_class,yan2023feature_one_class,sun2023hierarchical_one_class} mitigate labeling requirements but often generalize poorly across diverse scenes, leading to retraining and adaptation overhead. In contrast, VLM-based methods move beyond fixed-label recognition by enabling open-ended comprehension and allow anomalies to be flexibly defined via natural language. Building on these strengths, \name{} further employs a carefully designed pipeline to integrate pretrained multimodal knowledge with scene-specific rules, achieving stronger adaptability across environments without costly retraining.

\noindent \textbf{Vision-Language Models for VAD}. 
Recent VLMs, including GPT-4o~\cite{achiam2023gpt4}, Gemini~\cite{team2023gemini}, and QwenVL~\cite{bai2023qwenvlversatilevisionlanguagemodel}, enable zero-shot reasoning, semantic understanding, and natural-language interaction. These capabilities have inspired VAD systems such as LAVAD~\cite{zanella2024lavad}, AnomalyRuler~\cite{yang2024anomlayruler}, VERA~\cite{ye2025vera}, Hawk~\cite{tang2024hawk}, and Sherlock~\cite{ma2025sherlock}. While achieving strong accuracy, these systems typically incur high latency and resource usage. In contrast, \name{} adopts a cascaded design that first filters routine frames with lightweight CLIP models before applying fine-grained VLM reasoning, thereby achieving real-time efficiency without sacrificing accuracy.

\noindent \textbf{Prompt Engineering for VAD}. 
Prompting strategies have been explored to better align VLMs with anomaly cues, including text prompts~\cite{chen2024promptenhanced,yang2024textprompt,pu2024learningprompt}, joint visual-text prompts~\cite{wang2025federatedprompt,wu2024weaklyprompt,wu2024vadclipprompt}, and VLM-driven prompting~\cite{ye2025vera,zhu2024llms_prompt,wu2024weakly_llm_prompt}. However, many require iterative tuning or heavy preprocessing, limiting streaming deployment. In contrast, \name{} employs a training-free \textit{motion mask prompting} that highlights foreground moving regions as anomaly cues, reducing background distraction and enabling efficient anomaly detection in complex scenes.

\section{Limitations and Future Work}
There are also limitations in the current design of \name{}, which suggest directions for future research. In particular, while the system incorporates a feedback-driven rule evolution module to refine and expand its rule set, it still struggles to adapt when the boundary between normal and abnormal behaviors undergoes a fundamental shift (e.g., \textit{``a restricted area becoming a public space''}). Such concept drift remains a long-standing challenge in machine learning and anomaly detection. A promising direction is to develop mechanisms that can autonomously detect and adapt to evolving contexts, potentially by leveraging continual learning, meta-learning, or cross-scene transfer strategies. Addressing this challenge could pave the way toward long-term, fully adaptive VAD systems that remain robust in dynamic and continuously changing real-world environments.

\section{Conclusion}
\label{sec: conclusion}
In this paper, we introduce \name{}, a real-time VAD system that addresses the computational efficiency challenges in VLM-based approaches. Through a two-stage cascaded architecture combining lightweight CLIP-based filtering with VLM reasoning, \name{} achieves a 151.79$\times$ speedup while maintaining 97.2\% detection accuracy. The system's core innovation shifts from explicit anomaly enumeration to rule-based deviation detection, learning scene-specific behavioral norms offline for real-time inference. Motion mask prompting guides model attention to motion-relevant regions, while rule evolution enables continuous adaptation through automated and user feedback. Extensive evaluation across four datasets demonstrates practical deployment viability with 72.25 fps throughput, establishing \name{} as a scalable solution for safety-critical video applications.

\bibliographystyle{ACM-Reference-Format}
\bibliography{main} 


\begin{thebibliography}{67}


\ifx \showCODEN    \undefined \def \showCODEN     #1{\unskip}     \fi
\ifx \showDOI      \undefined \def \showDOI       #1{#1}\fi
\ifx \showISBNx    \undefined \def \showISBNx     #1{\unskip}     \fi
\ifx \showISBNxiii \undefined \def \showISBNxiii  #1{\unskip}     \fi
\ifx \showISSN     \undefined \def \showISSN      #1{\unskip}     \fi
\ifx \showLCCN     \undefined \def \showLCCN      #1{\unskip}     \fi
\ifx \shownote     \undefined \def \shownote      #1{#1}          \fi
\ifx \showarticletitle \undefined \def \showarticletitle #1{#1}   \fi
\ifx \showURL      \undefined \def \showURL       {\relax}        \fi
\providecommand\bibfield[2]{#2}
\providecommand\bibinfo[2]{#2}
\providecommand\natexlab[1]{#1}
\providecommand\showeprint[2][]{arXiv:#2}

\bibitem[\protect\citeauthoryear{Aboah}{Aboah}{2021}]%
        {aboah2021vision_fully_supervised}
\bibfield{author}{\bibinfo{person}{Armstrong Aboah}.} \bibinfo{year}{2021}\natexlab{}.
\newblock \showarticletitle{A vision-based system for traffic anomaly detection using deep learning and decision trees}.
\newblock \bibinfo{journal}{{\em IEEE CVPRW\/}} (\bibinfo{year}{2021}).
\newblock


\bibitem[\protect\citeauthoryear{Achiam, Adler, Agarwal, Ahmad, Akkaya, Aleman, Almeida, Altenschmidt, Altman, Anadkat, et~al\mbox{.}}{Achiam et~al\mbox{.}}{2023}]%
        {achiam2023gpt4}
\bibfield{author}{\bibinfo{person}{Josh Achiam}, \bibinfo{person}{Steven Adler}, \bibinfo{person}{Sandhini Agarwal}, \bibinfo{person}{Lama Ahmad}, \bibinfo{person}{Ilge Akkaya}, \bibinfo{person}{Florencia~Leoni Aleman}, \bibinfo{person}{Diogo Almeida}, \bibinfo{person}{Janko Altenschmidt}, \bibinfo{person}{Sam Altman}, \bibinfo{person}{Shyamal Anadkat}, {et~al\mbox{.}}} \bibinfo{year}{2023}\natexlab{}.
\newblock \showarticletitle{Gpt-4 technical report}.
\newblock \bibinfo{journal}{{\em arXiv preprint arXiv:2303.08774\/}} (\bibinfo{year}{2023}).
\newblock


\bibitem[\protect\citeauthoryear{Acsintoae, Florescu, Georgescu, Mare, Sumedrea, Ionescu, Khan, and Shah}{Acsintoae et~al\mbox{.}}{2022}]%
        {acsintoae2022UBnormal}
\bibfield{author}{\bibinfo{person}{Andra Acsintoae}, \bibinfo{person}{Andrei Florescu}, \bibinfo{person}{Mariana-Iuliana Georgescu}, \bibinfo{person}{Tudor Mare}, \bibinfo{person}{Paul Sumedrea}, \bibinfo{person}{Radu~Tudor Ionescu}, \bibinfo{person}{Fahad~Shahbaz Khan}, {and} \bibinfo{person}{Mubarak Shah}.} \bibinfo{year}{2022}\natexlab{}.
\newblock \showarticletitle{Ubnormal: New benchmark for supervised open-set video anomaly detection}.
\newblock \bibinfo{journal}{{\em IEEE CVPR\/}} (\bibinfo{year}{2022}).
\newblock


\bibitem[\protect\citeauthoryear{Bae, Ko, Song, and Yun}{Bae et~al\mbox{.}}{2023}]%
        {bae2023fast}
\bibfield{author}{\bibinfo{person}{Sangmin Bae}, \bibinfo{person}{Jongwoo Ko}, \bibinfo{person}{Hwanjun Song}, {and} \bibinfo{person}{Se-Young Yun}.} \bibinfo{year}{2023}\natexlab{}.
\newblock \showarticletitle{Fast and robust early-exiting framework for autoregressive language models with synchronized parallel decoding}.
\newblock \bibinfo{journal}{{\em ACL EMNLP\/}} (\bibinfo{year}{2023}).
\newblock


\bibitem[\protect\citeauthoryear{Bai, Bai, Yang, Wang, Tan, Wang, Lin, Zhou, and Zhou}{Bai et~al\mbox{.}}{2025a}]%
        {bai2023qwenvlversatilevisionlanguagemodel}
\bibfield{author}{\bibinfo{person}{Jinze Bai}, \bibinfo{person}{Shuai Bai}, \bibinfo{person}{Shusheng Yang}, \bibinfo{person}{Shijie Wang}, \bibinfo{person}{Sinan Tan}, \bibinfo{person}{Peng Wang}, \bibinfo{person}{Junyang Lin}, \bibinfo{person}{Chang Zhou}, {and} \bibinfo{person}{Jingren Zhou}.} \bibinfo{year}{2025}\natexlab{a}.
\newblock \showarticletitle{Qwen-VL: A Versatile Vision-Language Model for Understanding, Localization, Text Reading, and Beyond}.
\newblock \bibinfo{journal}{{\em arXiv preprint arXiv:2308.12966\/}} (\bibinfo{year}{2025}).
\newblock


\bibitem[\protect\citeauthoryear{Bai, Chen, Liu, Wang, Ge, Song, Dang, Wang, Wang, Tang, et~al\mbox{.}}{Bai et~al\mbox{.}}{2025b}]%
        {wang2024qwen2vl}
\bibfield{author}{\bibinfo{person}{Shuai Bai}, \bibinfo{person}{Keqin Chen}, \bibinfo{person}{Xuejing Liu}, \bibinfo{person}{Jialin Wang}, \bibinfo{person}{Wenbin Ge}, \bibinfo{person}{Sibo Song}, \bibinfo{person}{Kai Dang}, \bibinfo{person}{Peng Wang}, \bibinfo{person}{Shijie Wang}, \bibinfo{person}{Jun Tang}, {et~al\mbox{.}}} \bibinfo{year}{2025}\natexlab{b}.
\newblock \showarticletitle{Qwen2. 5-vl technical report}.
\newblock \bibinfo{journal}{{\em arXiv preprint arXiv:2502.13923\/}} (\bibinfo{year}{2025}).
\newblock


\bibitem[\protect\citeauthoryear{Bhardwaj, Xia, Ananthanarayanan, Jiang, Shu, Karianakis, Hsieh, Bahl, and Stoica}{Bhardwaj et~al\mbox{.}}{2022}]%
        {bhardwaj2022ekya}
\bibfield{author}{\bibinfo{person}{Romil Bhardwaj}, \bibinfo{person}{Zhengxu Xia}, \bibinfo{person}{Ganesh Ananthanarayanan}, \bibinfo{person}{Junchen Jiang}, \bibinfo{person}{Yuanchao Shu}, \bibinfo{person}{Nikolaos Karianakis}, \bibinfo{person}{Kevin Hsieh}, \bibinfo{person}{Paramvir Bahl}, {and} \bibinfo{person}{Ion Stoica}.} \bibinfo{year}{2022}\natexlab{}.
\newblock \showarticletitle{Ekya: Continuous learning of video analytics models on edge compute servers}.
\newblock \bibinfo{journal}{{\em USENIX NSDI\/}} (\bibinfo{year}{2022}).
\newblock


\bibitem[\protect\citeauthoryear{Bolya, Huang, Sun, Cho, Madotto, Wei, Ma, Zhi, Rajasegaran, Rasheed, et~al\mbox{.}}{Bolya et~al\mbox{.}}{2025}]%
        {bolya2025perception_encoder}
\bibfield{author}{\bibinfo{person}{Daniel Bolya}, \bibinfo{person}{Po-Yao Huang}, \bibinfo{person}{Peize Sun}, \bibinfo{person}{Jang~Hyun Cho}, \bibinfo{person}{Andrea Madotto}, \bibinfo{person}{Chen Wei}, \bibinfo{person}{Tengyu Ma}, \bibinfo{person}{Jiale Zhi}, \bibinfo{person}{Jathushan Rajasegaran}, \bibinfo{person}{Hanoona Rasheed}, {et~al\mbox{.}}} \bibinfo{year}{2025}\natexlab{}.
\newblock \showarticletitle{Perception encoder: The best visual embeddings are not at the output of the network}.
\newblock \bibinfo{journal}{{\em arXiv preprint arXiv:2504.13181\/}} (\bibinfo{year}{2025}).
\newblock


\bibitem[\protect\citeauthoryear{Cao, Lu, Wang, and Zhang}{Cao et~al\mbox{.}}{2023}]%
        {cao2023NWPUCampus}
\bibfield{author}{\bibinfo{person}{Congqi Cao}, \bibinfo{person}{Yue Lu}, \bibinfo{person}{Peng Wang}, {and} \bibinfo{person}{Yanning Zhang}.} \bibinfo{year}{2023}\natexlab{}.
\newblock \showarticletitle{A new comprehensive benchmark for semi-supervised video anomaly detection and anticipation}.
\newblock \bibinfo{journal}{{\em IEEE CVPR\/}} (\bibinfo{year}{2023}).
\newblock


\bibitem[\protect\citeauthoryear{Carreira and Zisserman}{Carreira and Zisserman}{2017}]%
        {carreira2017quo_i3d}
\bibfield{author}{\bibinfo{person}{Joao Carreira} {and} \bibinfo{person}{Andrew Zisserman}.} \bibinfo{year}{2017}\natexlab{}.
\newblock \showarticletitle{Quo vadis, action recognition? a new model and the kinetics dataset}.
\newblock \bibinfo{journal}{{\em proceedings of the IEEE Conference on Computer Vision and Pattern Recognition\/}} (\bibinfo{year}{2017}).
\newblock


\bibitem[\protect\citeauthoryear{Chen, Li, Su, Zha, and Huang}{Chen et~al\mbox{.}}{2024}]%
        {chen2024promptenhanced}
\bibfield{author}{\bibinfo{person}{Junxi Chen}, \bibinfo{person}{Liang Li}, \bibinfo{person}{Li Su}, \bibinfo{person}{Zheng-jun Zha}, {and} \bibinfo{person}{Qingming Huang}.} \bibinfo{year}{2024}\natexlab{}.
\newblock \showarticletitle{Prompt-enhanced multiple instance learning for weakly supervised video anomaly detection}.
\newblock \bibinfo{journal}{{\em IEEE CVPR\/}} (\bibinfo{year}{2024}).
\newblock


\bibitem[\protect\citeauthoryear{Ding, Zhang, Wu, Pang, Yang, Wang, and Zhang}{Ding et~al\mbox{.}}{2025}]%
        {ding2025slowfastvad}
\bibfield{author}{\bibinfo{person}{Zongcan Ding}, \bibinfo{person}{Haodong Zhang}, \bibinfo{person}{Peng Wu}, \bibinfo{person}{Guansong Pang}, \bibinfo{person}{Zhiwei Yang}, \bibinfo{person}{Peng Wang}, {and} \bibinfo{person}{Yanning Zhang}.} \bibinfo{year}{2025}\natexlab{}.
\newblock \showarticletitle{SlowFastVAD: Video Anomaly Detection via Integrating Simple Detector and RAG-Enhanced Vision-Language Model}.
\newblock \bibinfo{journal}{{\em arXiv preprint arXiv:2504.10320\/}} (\bibinfo{year}{2025}).
\newblock


\bibitem[\protect\citeauthoryear{Feng, Li, Lin, and Chen}{Feng et~al\mbox{.}}{2025}]%
        {feng2025align}
\bibfield{author}{\bibinfo{person}{Qianhan Feng}, \bibinfo{person}{Wenshuo Li}, \bibinfo{person}{Tong Lin}, {and} \bibinfo{person}{Xinghao Chen}.} \bibinfo{year}{2025}\natexlab{}.
\newblock \showarticletitle{Align-KD: Distilling Cross-Modal Alignment Knowledge for Mobile Vision-Language Large Model Enhancement}.
\newblock \bibinfo{journal}{{\em Proceedings of the Computer Vision and Pattern Recognition Conference\/}} (\bibinfo{year}{2025}).
\newblock


\bibitem[\protect\citeauthoryear{Guo, Yang, Zhang, Song, Zhang, Xu, Zhu, Ma, Wang, Bi, et~al\mbox{.}}{Guo et~al\mbox{.}}{2025}]%
        {guo2025deepseek_r1}
\bibfield{author}{\bibinfo{person}{Daya Guo}, \bibinfo{person}{Dejian Yang}, \bibinfo{person}{Haowei Zhang}, \bibinfo{person}{Junxiao Song}, \bibinfo{person}{Ruoyu Zhang}, \bibinfo{person}{Runxin Xu}, \bibinfo{person}{Qihao Zhu}, \bibinfo{person}{Shirong Ma}, \bibinfo{person}{Peiyi Wang}, \bibinfo{person}{Xiao Bi}, {et~al\mbox{.}}} \bibinfo{year}{2025}\natexlab{}.
\newblock \showarticletitle{Deepseek-r1: Incentivizing reasoning capability in llms via reinforcement learning}.
\newblock \bibinfo{journal}{{\em arXiv preprint arXiv:2501.12948\/}} (\bibinfo{year}{2025}).
\newblock


\bibitem[\protect\citeauthoryear{Jiang, Ananthanarayanan, Bodik, Sen, and Stoica}{Jiang et~al\mbox{.}}{2018}]%
        {jiang2018chameleon}
\bibfield{author}{\bibinfo{person}{Junchen Jiang}, \bibinfo{person}{Ganesh Ananthanarayanan}, \bibinfo{person}{Peter Bodik}, \bibinfo{person}{Siddhartha Sen}, {and} \bibinfo{person}{Ion Stoica}.} \bibinfo{year}{2018}\natexlab{}.
\newblock \showarticletitle{Chameleon: scalable adaptation of video analytics}.
\newblock \bibinfo{journal}{{\em ACM SIGCOMM\/}} (\bibinfo{year}{2018}).
\newblock


\bibitem[\protect\citeauthoryear{Jiang, Lin, Li, Shu, and Liu}{Jiang et~al\mbox{.}}{2021}]%
        {jiang2021flexible}
\bibfield{author}{\bibinfo{person}{Shiqi Jiang}, \bibinfo{person}{Zhiqi Lin}, \bibinfo{person}{Yuanchun Li}, \bibinfo{person}{Yuanchao Shu}, {and} \bibinfo{person}{Yunxin Liu}.} \bibinfo{year}{2021}\natexlab{}.
\newblock \showarticletitle{Flexible high-resolution object detection on edge devices with tunable latency}.
\newblock \bibinfo{journal}{{\em ACM MobiCom\/}} (\bibinfo{year}{2021}).
\newblock


\bibitem[\protect\citeauthoryear{Li, Xu, Tian, Wang, Yan, Bi, Ye, Chen, Xu, Cao, et~al\mbox{.}}{Li et~al\mbox{.}}{2022c}]%
        {li2022mplug}
\bibfield{author}{\bibinfo{person}{Chenliang Li}, \bibinfo{person}{Haiyang Xu}, \bibinfo{person}{Junfeng Tian}, \bibinfo{person}{Wei Wang}, \bibinfo{person}{Ming Yan}, \bibinfo{person}{Bin Bi}, \bibinfo{person}{Jiabo Ye}, \bibinfo{person}{Hehong Chen}, \bibinfo{person}{Guohai Xu}, \bibinfo{person}{Zheng Cao}, {et~al\mbox{.}}} \bibinfo{year}{2022}\natexlab{c}.
\newblock \showarticletitle{mplug: Effective and efficient vision-language learning by cross-modal skip-connections}.
\newblock \bibinfo{journal}{{\em ACL EMNLP\/}} (\bibinfo{year}{2022}).
\newblock


\bibitem[\protect\citeauthoryear{Li, Cai, Zeng, and Zhao}{Li et~al\mbox{.}}{2022a}]%
        {li2022scale_weakly_supervised}
\bibfield{author}{\bibinfo{person}{Guoqiu Li}, \bibinfo{person}{Guanxiong Cai}, \bibinfo{person}{Xingyu Zeng}, {and} \bibinfo{person}{Rui Zhao}.} \bibinfo{year}{2022}\natexlab{a}.
\newblock \showarticletitle{Scale-aware spatio-temporal relation learning for video anomaly detection}.
\newblock \bibinfo{journal}{{\em Springer ECCV\/}} (\bibinfo{year}{2022}).
\newblock


\bibitem[\protect\citeauthoryear{Li, Liu, and Jiao}{Li et~al\mbox{.}}{2022b}]%
        {li2022self_weakly_supervised}
\bibfield{author}{\bibinfo{person}{Shuo Li}, \bibinfo{person}{Fang Liu}, {and} \bibinfo{person}{Licheng Jiao}.} \bibinfo{year}{2022}\natexlab{b}.
\newblock \showarticletitle{Self-training multi-sequence learning with transformer for weakly supervised video anomaly detection}.
\newblock \bibinfo{journal}{{\em AAAI\/}} (\bibinfo{year}{2022}).
\newblock


\bibitem[\protect\citeauthoryear{Li, Chen, Hu, Wang, Shi, and Zhang}{Li et~al\mbox{.}}{2024}]%
        {li2024videovista}
\bibfield{author}{\bibinfo{person}{Yunxin Li}, \bibinfo{person}{Xinyu Chen}, \bibinfo{person}{Baotian Hu}, \bibinfo{person}{Longyue Wang}, \bibinfo{person}{Haoyuan Shi}, {and} \bibinfo{person}{Min Zhang}.} \bibinfo{year}{2024}\natexlab{}.
\newblock \showarticletitle{Videovista: A versatile benchmark for video understanding and reasoning}.
\newblock \bibinfo{journal}{{\em arXiv preprint arXiv:2406.11303\/}} (\bibinfo{year}{2024}).
\newblock


\bibitem[\protect\citeauthoryear{Liu, Liu, Lin, Li, Cao, Sun, Hu, Song, Boukerche, and Leung}{Liu et~al\mbox{.}}{2025}]%
        {liu2025network_system_vad}
\bibfield{author}{\bibinfo{person}{Jing Liu}, \bibinfo{person}{Yang Liu}, \bibinfo{person}{Jieyu Lin}, \bibinfo{person}{Jielin Li}, \bibinfo{person}{Liang Cao}, \bibinfo{person}{Peng Sun}, \bibinfo{person}{Bo Hu}, \bibinfo{person}{Liang Song}, \bibinfo{person}{Azzedine Boukerche}, {and} \bibinfo{person}{Victor~CM Leung}.} \bibinfo{year}{2025}\natexlab{}.
\newblock \showarticletitle{Networking systems for video anomaly detection: A tutorial and survey}.
\newblock \bibinfo{journal}{{\it Comput. Surveys}} (\bibinfo{year}{2025}).
\newblock


\bibitem[\protect\citeauthoryear{Liu, Zhang, Lyu, Zhang, Xiao, Shen, and Yu}{Liu et~al\mbox{.}}{2022}]%
        {liu2022traffic_fully_supervised}
\bibfield{author}{\bibinfo{person}{Xiaoming Liu}, \bibinfo{person}{Zhanwei Zhang}, \bibinfo{person}{Lingjuan Lyu}, \bibinfo{person}{Zhaohan Zhang}, \bibinfo{person}{Shuai Xiao}, \bibinfo{person}{Chao Shen}, {and} \bibinfo{person}{Philip~S Yu}.} \bibinfo{year}{2022}\natexlab{}.
\newblock \showarticletitle{Traffic anomaly prediction based on joint static-dynamic spatio-temporal evolutionary learning}.
\newblock \bibinfo{journal}{{\em IEEE Transactions on Knowledge and Data Engineering\/}} (\bibinfo{year}{2022}).
\newblock


\bibitem[\protect\citeauthoryear{Liu, Nie, Long, Zhang, and Li}{Liu et~al\mbox{.}}{2021}]%
        {liu2021hybrid_one_class}
\bibfield{author}{\bibinfo{person}{Zhian Liu}, \bibinfo{person}{Yongwei Nie}, \bibinfo{person}{Chengjiang Long}, \bibinfo{person}{Qing Zhang}, {and} \bibinfo{person}{Guiqing Li}.} \bibinfo{year}{2021}\natexlab{}.
\newblock \showarticletitle{A hybrid video anomaly detection framework via memory-augmented flow reconstruction and flow-guided frame prediction}.
\newblock \bibinfo{journal}{{\em IEEE ICCV\/}} (\bibinfo{year}{2021}).
\newblock


\bibitem[\protect\citeauthoryear{Lu, Shi, and Jia}{Lu et~al\mbox{.}}{2013}]%
        {lu2013Avenue}
\bibfield{author}{\bibinfo{person}{Cewu Lu}, \bibinfo{person}{Jianping Shi}, {and} \bibinfo{person}{Jiaya Jia}.} \bibinfo{year}{2013}\natexlab{}.
\newblock \showarticletitle{Abnormal event detection at 150 fps in matlab}.
\newblock \bibinfo{journal}{{\em IEEE ICCV\/}} (\bibinfo{year}{2013}).
\newblock


\bibitem[\protect\citeauthoryear{Luo, Liu, and Gao}{Luo et~al\mbox{.}}{2017}]%
        {luo2017SHTech}
\bibfield{author}{\bibinfo{person}{Weixin Luo}, \bibinfo{person}{Wen Liu}, {and} \bibinfo{person}{Shenghua Gao}.} \bibinfo{year}{2017}\natexlab{}.
\newblock \showarticletitle{A revisit of sparse coding based anomaly detection in stacked rnn framework}.
\newblock \bibinfo{journal}{{\em IEEE ICCV\/}} (\bibinfo{year}{2017}).
\newblock


\bibitem[\protect\citeauthoryear{Ma, Wang, Luo, Yu, and Zhou}{Ma et~al\mbox{.}}{2025}]%
        {ma2025sherlock}
\bibfield{author}{\bibinfo{person}{Junxiao Ma}, \bibinfo{person}{Jingjing Wang}, \bibinfo{person}{Jiamin Luo}, \bibinfo{person}{Peiying Yu}, {and} \bibinfo{person}{Guodong Zhou}.} \bibinfo{year}{2025}\natexlab{}.
\newblock \showarticletitle{Sherlock: Towards Multi-scene Video Abnormal Event Extraction and Localization via a Global-local Spatial-sensitive LLM}.
\newblock \bibinfo{journal}{{\em ACM WWW\/}} (\bibinfo{year}{2025}).
\newblock


\bibitem[\protect\citeauthoryear{Noghre, Pazho, and Tabkhi}{Noghre et~al\mbox{.}}{2024}]%
        {noghre2024exploratory}
\bibfield{author}{\bibinfo{person}{Ghazal~Alinezhad Noghre}, \bibinfo{person}{Armin~Danesh Pazho}, {and} \bibinfo{person}{Hamed Tabkhi}.} \bibinfo{year}{2024}\natexlab{}.
\newblock \showarticletitle{An exploratory study on human-centric video anomaly detection through variational autoencoders and trajectory prediction}.
\newblock \bibinfo{journal}{{\em IEEE WACV\/}} (\bibinfo{year}{2024}).
\newblock


\bibitem[\protect\citeauthoryear{NVIDIA}{NVIDIA}{2025}]%
        {nvidial40s}
\bibfield{author}{\bibinfo{person}{NVIDIA}.} \bibinfo{year}{2025}\natexlab{}.
\newblock \bibinfo{title}{NVIDIA L40S}.
\newblock \bibinfo{howpublished}{\url{https://www.nvidia.com/en-us/data-center/l40s/}}.   (\bibinfo{year}{2025}).
\newblock
\newblock
\shownote{Accessed on June 23, 2025.}


\bibitem[\protect\citeauthoryear{OpenCV}{OpenCV}{2013}]%
        {opencv}
\bibfield{author}{\bibinfo{person}{OpenCV}.} \bibinfo{year}{2013}\natexlab{}.
\newblock \bibinfo{title}{OpenCV: Open Source Computer Vision Library}.
\newblock \bibinfo{howpublished}{\url{https://github.com/opencv/opencv}}.   (\bibinfo{year}{2013}).
\newblock
\newblock
\shownote{Accessed on June 26, 2025.}


\bibitem[\protect\citeauthoryear{Pu, Wu, Yang, and Wang}{Pu et~al\mbox{.}}{2024}]%
        {pu2024learningprompt}
\bibfield{author}{\bibinfo{person}{Yujiang Pu}, \bibinfo{person}{Xiaoyu Wu}, \bibinfo{person}{Lulu Yang}, {and} \bibinfo{person}{Shengjin Wang}.} \bibinfo{year}{2024}\natexlab{}.
\newblock \showarticletitle{Learning prompt-enhanced context features for weakly-supervised video anomaly detection}.
\newblock \bibinfo{journal}{{\em IEEE Transactions on Image Processing\/}} (\bibinfo{year}{2024}).
\newblock


\bibitem[\protect\citeauthoryear{Radford, Kim, Hallacy, Ramesh, Goh, Agarwal, Sastry, Askell, Mishkin, Clark, et~al\mbox{.}}{Radford et~al\mbox{.}}{2021}]%
        {radford2021learning_CLIP}
\bibfield{author}{\bibinfo{person}{Alec Radford}, \bibinfo{person}{Jong~Wook Kim}, \bibinfo{person}{Chris Hallacy}, \bibinfo{person}{Aditya Ramesh}, \bibinfo{person}{Gabriel Goh}, \bibinfo{person}{Sandhini Agarwal}, \bibinfo{person}{Girish Sastry}, \bibinfo{person}{Amanda Askell}, \bibinfo{person}{Pamela Mishkin}, \bibinfo{person}{Jack Clark}, {et~al\mbox{.}}} \bibinfo{year}{2021}\natexlab{}.
\newblock \showarticletitle{Learning transferable visual models from natural language supervision}.
\newblock \bibinfo{journal}{{\em ICML\/}} (\bibinfo{year}{2021}).
\newblock


\bibitem[\protect\citeauthoryear{Ramachandra, Jones, and Vatsavai}{Ramachandra et~al\mbox{.}}{2020}]%
        {ramachandra2020survey}
\bibfield{author}{\bibinfo{person}{Bharathkumar Ramachandra}, \bibinfo{person}{Michael~J Jones}, {and} \bibinfo{person}{Ranga~Raju Vatsavai}.} \bibinfo{year}{2020}\natexlab{}.
\newblock \showarticletitle{A survey of single-scene video anomaly detection}.
\newblock \bibinfo{journal}{{\em IEEE transactions on pattern analysis and machine intelligence\/}} (\bibinfo{year}{2020}).
\newblock


\bibitem[\protect\citeauthoryear{Reich and Schultz}{Reich and Schultz}{2024}]%
        {reich2024uncovering}
\bibfield{author}{\bibinfo{person}{Daniel Reich} {and} \bibinfo{person}{Tanja Schultz}.} \bibinfo{year}{2024}\natexlab{}.
\newblock \showarticletitle{Uncovering the Full Potential of Visual Grounding Methods in VQA}.
\newblock \bibinfo{journal}{{\em ACL ACL\/}} (\bibinfo{year}{2024}).
\newblock


\bibitem[\protect\citeauthoryear{Shtedritski, Rupprecht, and Vedaldi}{Shtedritski et~al\mbox{.}}{2023}]%
        {shtedritski2023does_red_circle}
\bibfield{author}{\bibinfo{person}{Aleksandar Shtedritski}, \bibinfo{person}{Christian Rupprecht}, {and} \bibinfo{person}{Andrea Vedaldi}.} \bibinfo{year}{2023}\natexlab{}.
\newblock \showarticletitle{What does clip know about a red circle? visual prompt engineering for vlms}.
\newblock \bibinfo{journal}{{\em IEEE ICCV\/}} (\bibinfo{year}{2023}).
\newblock


\bibitem[\protect\citeauthoryear{Sultani, Chen, and Shah}{Sultani et~al\mbox{.}}{2018a}]%
        {sultani2018realworld}
\bibfield{author}{\bibinfo{person}{Waqas Sultani}, \bibinfo{person}{Chen Chen}, {and} \bibinfo{person}{Mubarak Shah}.} \bibinfo{year}{2018}\natexlab{a}.
\newblock \showarticletitle{Real-world anomaly detection in surveillance videos}.
\newblock \bibinfo{journal}{{\em IEEE CVPR\/}} (\bibinfo{year}{2018}).
\newblock


\bibitem[\protect\citeauthoryear{Sultani, Chen, and Shah}{Sultani et~al\mbox{.}}{2018b}]%
        {sultani2018real_UCF_Crime}
\bibfield{author}{\bibinfo{person}{Waqas Sultani}, \bibinfo{person}{Chen Chen}, {and} \bibinfo{person}{Mubarak Shah}.} \bibinfo{year}{2018}\natexlab{b}.
\newblock \showarticletitle{Real-world anomaly detection in surveillance videos}.
\newblock \bibinfo{journal}{{\em IEEE CVPR\/}} (\bibinfo{year}{2018}).
\newblock


\bibitem[\protect\citeauthoryear{Sun and Gong}{Sun and Gong}{2023}]%
        {sun2023hierarchical_one_class}
\bibfield{author}{\bibinfo{person}{Shengyang Sun} {and} \bibinfo{person}{Xiaojin Gong}.} \bibinfo{year}{2023}\natexlab{}.
\newblock \showarticletitle{Hierarchical semantic contrast for scene-aware video anomaly detection}.
\newblock \bibinfo{journal}{{\em IEEE CVPR\/}} (\bibinfo{year}{2023}).
\newblock


\bibitem[\protect\citeauthoryear{Tang, Lu, Wu, Xu, Ma, Fang, Guo, Lu, Chen, and Chen}{Tang et~al\mbox{.}}{2024}]%
        {tang2024hawk}
\bibfield{author}{\bibinfo{person}{Jiaqi Tang}, \bibinfo{person}{Hao Lu}, \bibinfo{person}{Ruizheng Wu}, \bibinfo{person}{Xiaogang Xu}, \bibinfo{person}{Ke Ma}, \bibinfo{person}{Cheng Fang}, \bibinfo{person}{Bin Guo}, \bibinfo{person}{Jiangbo Lu}, \bibinfo{person}{Qifeng Chen}, {and} \bibinfo{person}{Yingcong Chen}.} \bibinfo{year}{2024}\natexlab{}.
\newblock \showarticletitle{Hawk: Learning to understand open-world video anomalies}.
\newblock \bibinfo{journal}{{\em NeurIPS\/}} (\bibinfo{year}{2024}).
\newblock


\bibitem[\protect\citeauthoryear{Team, Anil, Borgeaud, Alayrac, Yu, Soricut, Schalkwyk, Dai, Hauth, Millican, et~al\mbox{.}}{Team et~al\mbox{.}}{2023}]%
        {team2023gemini}
\bibfield{author}{\bibinfo{person}{Gemini Team}, \bibinfo{person}{Rohan Anil}, \bibinfo{person}{Sebastian Borgeaud}, \bibinfo{person}{Jean-Baptiste Alayrac}, \bibinfo{person}{Jiahui Yu}, \bibinfo{person}{Radu Soricut}, \bibinfo{person}{Johan Schalkwyk}, \bibinfo{person}{Andrew~M Dai}, \bibinfo{person}{Anja Hauth}, \bibinfo{person}{Katie Millican}, {et~al\mbox{.}}} \bibinfo{year}{2023}\natexlab{}.
\newblock \showarticletitle{Gemini: a family of highly capable multimodal models}.
\newblock \bibinfo{journal}{{\em arXiv preprint arXiv:2312.11805\/}} (\bibinfo{year}{2023}).
\newblock


\bibitem[\protect\citeauthoryear{Thakare, Dogra, Choi, Kim, and Kim}{Thakare et~al\mbox{.}}{2023}]%
        {thakare2023rareanom}
\bibfield{author}{\bibinfo{person}{Kamalakar~Vijay Thakare}, \bibinfo{person}{Debi~Prosad Dogra}, \bibinfo{person}{Heeseung Choi}, \bibinfo{person}{Haksub Kim}, {and} \bibinfo{person}{Ig-Jae Kim}.} \bibinfo{year}{2023}\natexlab{}.
\newblock \showarticletitle{Rareanom: A benchmark video dataset for rare type anomalies}.
\newblock \bibinfo{journal}{{\em Pattern Recognition\/}} (\bibinfo{year}{2023}).
\newblock


\bibitem[\protect\citeauthoryear{Tian, Pang, Chen, Singh, Verjans, and Carneiro}{Tian et~al\mbox{.}}{2021}]%
        {tian2021weakly_weakly_supervised}
\bibfield{author}{\bibinfo{person}{Yu Tian}, \bibinfo{person}{Guansong Pang}, \bibinfo{person}{Yuanhong Chen}, \bibinfo{person}{Rajvinder Singh}, \bibinfo{person}{Johan~W Verjans}, {and} \bibinfo{person}{Gustavo Carneiro}.} \bibinfo{year}{2021}\natexlab{}.
\newblock \showarticletitle{Weakly-supervised video anomaly detection with robust temporal feature magnitude learning}.
\newblock \bibinfo{journal}{{\em IEEE ICCV\/}} (\bibinfo{year}{2021}).
\newblock


\bibitem[\protect\citeauthoryear{Vaswani, Shazeer, Parmar, Uszkoreit, Jones, Gomez, Kaiser, and Polosukhin}{Vaswani et~al\mbox{.}}{2017}]%
        {vaswani2017attention}
\bibfield{author}{\bibinfo{person}{Ashish Vaswani}, \bibinfo{person}{Noam Shazeer}, \bibinfo{person}{Niki Parmar}, \bibinfo{person}{Jakob Uszkoreit}, \bibinfo{person}{Llion Jones}, \bibinfo{person}{Aidan~N Gomez}, \bibinfo{person}{{\L}ukasz Kaiser}, {and} \bibinfo{person}{Illia Polosukhin}.} \bibinfo{year}{2017}\natexlab{}.
\newblock \showarticletitle{Attention is all you need}.
\newblock \bibinfo{journal}{{\em NeurIPS\/}} (\bibinfo{year}{2017}).
\newblock


\bibitem[\protect\citeauthoryear{Wang, Chen, Liu, Chen, Lin, Han, et~al\mbox{.}}{Wang et~al\mbox{.}}{2024}]%
        {wang2024yolov10}
\bibfield{author}{\bibinfo{person}{Ao Wang}, \bibinfo{person}{Hui Chen}, \bibinfo{person}{Lihao Liu}, \bibinfo{person}{Kai Chen}, \bibinfo{person}{Zijia Lin}, \bibinfo{person}{Jungong Han}, {et~al\mbox{.}}} \bibinfo{year}{2024}\natexlab{}.
\newblock \showarticletitle{Yolov10: Real-time end-to-end object detection}.
\newblock \bibinfo{journal}{{\em NeurIPS\/}} (\bibinfo{year}{2024}).
\newblock


\bibitem[\protect\citeauthoryear{Wang, Huang, Wen, Wang, Liu, and Xu}{Wang et~al\mbox{.}}{2025}]%
        {wang2025federatedprompt}
\bibfield{author}{\bibinfo{person}{Benfeng Wang}, \bibinfo{person}{Chao Huang}, \bibinfo{person}{Jie Wen}, \bibinfo{person}{Wei Wang}, \bibinfo{person}{Yabo Liu}, {and} \bibinfo{person}{Yong Xu}.} \bibinfo{year}{2025}\natexlab{}.
\newblock \showarticletitle{Federated Weakly Supervised Video Anomaly Detection with Multimodal Prompt}.
\newblock \bibinfo{journal}{{\em AAAI\/}} (\bibinfo{year}{2025}).
\newblock


\bibitem[\protect\citeauthoryear{Wang and Cherian}{Wang and Cherian}{2019}]%
        {wang2019gods}
\bibfield{author}{\bibinfo{person}{Jue Wang} {and} \bibinfo{person}{Anoop Cherian}.} \bibinfo{year}{2019}\natexlab{}.
\newblock \showarticletitle{Gods: Generalized one-class discriminative subspaces for anomaly detection}. In \bibinfo{booktitle}{{\em Proceedings of the IEEE/CVF International Conference on Computer Vision}}. \bibinfo{pages}{8201--8211}.
\newblock


\bibitem[\protect\citeauthoryear{Wang, Che, Jiang, Xiao, Yang, Tang, Ye, Wang, and Qi}{Wang et~al\mbox{.}}{2021}]%
        {wang2021robust_unsupervised}
\bibfield{author}{\bibinfo{person}{Xuanzhao Wang}, \bibinfo{person}{Zhengping Che}, \bibinfo{person}{Bo Jiang}, \bibinfo{person}{Ning Xiao}, \bibinfo{person}{Ke Yang}, \bibinfo{person}{Jian Tang}, \bibinfo{person}{Jieping Ye}, \bibinfo{person}{Jingyu Wang}, {and} \bibinfo{person}{Qi Qi}.} \bibinfo{year}{2021}\natexlab{}.
\newblock \showarticletitle{Robust unsupervised video anomaly detection by multipath frame prediction}.
\newblock \bibinfo{journal}{{\em IEEE transactions on neural networks and learning systems\/}} (\bibinfo{year}{2021}).
\newblock


\bibitem[\protect\citeauthoryear{Wu, Hsieh, Chen, Fuh, and Liu}{Wu et~al\mbox{.}}{2022}]%
        {wu2022self_unsupervised}
\bibfield{author}{\bibinfo{person}{Jhih-Ciang Wu}, \bibinfo{person}{He-Yen Hsieh}, \bibinfo{person}{Ding-Jie Chen}, \bibinfo{person}{Chiou-Shann Fuh}, {and} \bibinfo{person}{Tyng-Luh Liu}.} \bibinfo{year}{2022}\natexlab{}.
\newblock \showarticletitle{Self-supervised sparse representation for video anomaly detection}.
\newblock \bibinfo{journal}{{\em Springer ECCV\/}} (\bibinfo{year}{2022}).
\newblock


\bibitem[\protect\citeauthoryear{Wu, Liu, Shi, Sun, Shao, Wu, and Yang}{Wu et~al\mbox{.}}{2020}]%
        {wu2020not_XD_Violence}
\bibfield{author}{\bibinfo{person}{Peng Wu}, \bibinfo{person}{Jing Liu}, \bibinfo{person}{Yujia Shi}, \bibinfo{person}{Yujia Sun}, \bibinfo{person}{Fangtao Shao}, \bibinfo{person}{Zhaoyang Wu}, {and} \bibinfo{person}{Zhiwei Yang}.} \bibinfo{year}{2020}\natexlab{}.
\newblock \showarticletitle{Not only look, but also listen: Learning multimodal violence detection under weak supervision}.
\newblock \bibinfo{journal}{{\em Springer ECCV\/}} (\bibinfo{year}{2020}).
\newblock


\bibitem[\protect\citeauthoryear{Wu, Zhou, Pang, Yang, Yan, Wang, and Zhang}{Wu et~al\mbox{.}}{2024a}]%
        {wu2024weaklyprompt}
\bibfield{author}{\bibinfo{person}{Peng Wu}, \bibinfo{person}{Xuerong Zhou}, \bibinfo{person}{Guansong Pang}, \bibinfo{person}{Zhiwei Yang}, \bibinfo{person}{Qingsen Yan}, \bibinfo{person}{Peng Wang}, {and} \bibinfo{person}{Yanning Zhang}.} \bibinfo{year}{2024}\natexlab{a}.
\newblock \showarticletitle{Weakly supervised video anomaly detection and localization with spatio-temporal prompts}.
\newblock \bibinfo{journal}{{\em ACM MM\/}} (\bibinfo{year}{2024}).
\newblock


\bibitem[\protect\citeauthoryear{Wu, Zhou, Pang, Yang, Yan, Wang, and Zhang}{Wu et~al\mbox{.}}{2024b}]%
        {wu2024weakly_llm_prompt}
\bibfield{author}{\bibinfo{person}{Peng Wu}, \bibinfo{person}{Xuerong Zhou}, \bibinfo{person}{Guansong Pang}, \bibinfo{person}{Zhiwei Yang}, \bibinfo{person}{Qingsen Yan}, \bibinfo{person}{Peng Wang}, {and} \bibinfo{person}{Yanning Zhang}.} \bibinfo{year}{2024}\natexlab{b}.
\newblock \showarticletitle{Weakly supervised video anomaly detection and localization with spatio-temporal prompts}.
\newblock \bibinfo{journal}{{\em ACM MM\/}} (\bibinfo{year}{2024}).
\newblock


\bibitem[\protect\citeauthoryear{Wu, Zhou, Pang, Zhou, Yan, Wang, and Zhang}{Wu et~al\mbox{.}}{2024c}]%
        {wu2024vadclipprompt}
\bibfield{author}{\bibinfo{person}{Peng Wu}, \bibinfo{person}{Xuerong Zhou}, \bibinfo{person}{Guansong Pang}, \bibinfo{person}{Lingru Zhou}, \bibinfo{person}{Qingsen Yan}, \bibinfo{person}{Peng Wang}, {and} \bibinfo{person}{Yanning Zhang}.} \bibinfo{year}{2024}\natexlab{c}.
\newblock \showarticletitle{Vadclip: Adapting vision-language models for weakly supervised video anomaly detection}.
\newblock \bibinfo{journal}{{\em AAAI\/}} (\bibinfo{year}{2024}).
\newblock


\bibitem[\protect\citeauthoryear{Xiao, Yao, Li, and Chua}{Xiao et~al\mbox{.}}{2024}]%
        {xiao2024can}
\bibfield{author}{\bibinfo{person}{Junbin Xiao}, \bibinfo{person}{Angela Yao}, \bibinfo{person}{Yicong Li}, {and} \bibinfo{person}{Tat-Seng Chua}.} \bibinfo{year}{2024}\natexlab{}.
\newblock \showarticletitle{Can i trust your answer? visually grounded video question answering}.
\newblock \bibinfo{journal}{{\em IEEE CVPR\/}} (\bibinfo{year}{2024}).
\newblock


\bibitem[\protect\citeauthoryear{Yan, Zhang, Liu, Pang, and Wang}{Yan et~al\mbox{.}}{2023}]%
        {yan2023feature_one_class}
\bibfield{author}{\bibinfo{person}{Cheng Yan}, \bibinfo{person}{Shiyu Zhang}, \bibinfo{person}{Yang Liu}, \bibinfo{person}{Guansong Pang}, {and} \bibinfo{person}{Wenjun Wang}.} \bibinfo{year}{2023}\natexlab{}.
\newblock \showarticletitle{Feature prediction diffusion model for video anomaly detection}.
\newblock \bibinfo{journal}{{\em IEEE ICCV\/}} (\bibinfo{year}{2023}).
\newblock


\bibitem[\protect\citeauthoryear{Yan, Jiang, Cao, Yang, Yang, Shu, Yang, and Qiu}{Yan et~al\mbox{.}}{2026}]%
        {yan2025empowering}
\bibfield{author}{\bibinfo{person}{Yuxuan Yan}, \bibinfo{person}{Shiqi Jiang}, \bibinfo{person}{Ting Cao}, \bibinfo{person}{Yifan Yang}, \bibinfo{person}{Qianqian Yang}, \bibinfo{person}{Yuanchao Shu}, \bibinfo{person}{Yuqing Yang}, {and} \bibinfo{person}{Lili Qiu}.} \bibinfo{year}{2026}\natexlab{}.
\newblock \showarticletitle{Empowering agentic video analytics systems with video language models}.
\newblock \bibinfo{journal}{{\em USENIX NSDI\/}} (\bibinfo{year}{2026}).
\newblock


\bibitem[\protect\citeauthoryear{Yang, Lee, Dariush, Cao, and Lo}{Yang et~al\mbox{.}}{2024a}]%
        {yang2024anomlayruler}
\bibfield{author}{\bibinfo{person}{Yuchen Yang}, \bibinfo{person}{Kwonjoon Lee}, \bibinfo{person}{Behzad Dariush}, \bibinfo{person}{Yinzhi Cao}, {and} \bibinfo{person}{Shao-Yuan Lo}.} \bibinfo{year}{2024}\natexlab{a}.
\newblock \showarticletitle{Follow the rules: reasoning for video anomaly detection with large language models}.
\newblock \bibinfo{journal}{{\em Springer ECCV\/}} (\bibinfo{year}{2024}).
\newblock


\bibitem[\protect\citeauthoryear{Yang, Liu, and Wu}{Yang et~al\mbox{.}}{2024b}]%
        {yang2024textprompt}
\bibfield{author}{\bibinfo{person}{Zhiwei Yang}, \bibinfo{person}{Jing Liu}, {and} \bibinfo{person}{Peng Wu}.} \bibinfo{year}{2024}\natexlab{b}.
\newblock \showarticletitle{Text prompt with normality guidance for weakly supervised video anomaly detection}.
\newblock \bibinfo{journal}{{\em IEEE CVPR\/}} (\bibinfo{year}{2024}).
\newblock


\bibitem[\protect\citeauthoryear{Ye, Liu, and He}{Ye et~al\mbox{.}}{2025}]%
        {ye2025vera}
\bibfield{author}{\bibinfo{person}{Muchao Ye}, \bibinfo{person}{Weiyang Liu}, {and} \bibinfo{person}{Pan He}.} \bibinfo{year}{2025}\natexlab{}.
\newblock \showarticletitle{Vera: Explainable video anomaly detection via verbalized learning of vision-language models}.
\newblock \bibinfo{journal}{{\em IEEE CVPR\/}} (\bibinfo{year}{2025}).
\newblock


\bibitem[\protect\citeauthoryear{You, Fu, Wang, Yazdanbakhsh, and Lin}{You et~al\mbox{.}}{2024}]%
        {you2024linear}
\bibfield{author}{\bibinfo{person}{Haoran You}, \bibinfo{person}{Yichao Fu}, \bibinfo{person}{Zheng Wang}, \bibinfo{person}{Amir Yazdanbakhsh}, {and} \bibinfo{person}{Yingyan~Celine Lin}.} \bibinfo{year}{2024}\natexlab{}.
\newblock \showarticletitle{When linear attention meets autoregressive decoding: Towards more effective and efficient linearized large language models}.
\newblock \bibinfo{journal}{{\em ACM ICML\/}} (\bibinfo{year}{2024}).
\newblock


\bibitem[\protect\citeauthoryear{Zaheer, Mahmood, Khan, Segu, Yu, and Lee}{Zaheer et~al\mbox{.}}{2022}]%
        {zaheer2022generative_unsupervised}
\bibfield{author}{\bibinfo{person}{M~Zaigham Zaheer}, \bibinfo{person}{Arif Mahmood}, \bibinfo{person}{M~Haris Khan}, \bibinfo{person}{Mattia Segu}, \bibinfo{person}{Fisher Yu}, {and} \bibinfo{person}{Seung-Ik Lee}.} \bibinfo{year}{2022}\natexlab{}.
\newblock \showarticletitle{Generative cooperative learning for unsupervised video anomaly detection}.
\newblock \bibinfo{journal}{{\em IEEE CVPR\/}} (\bibinfo{year}{2022}).
\newblock


\bibitem[\protect\citeauthoryear{Zanella, Menapace, Mancini, Wang, and Ricci}{Zanella et~al\mbox{.}}{2024}]%
        {zanella2024lavad}
\bibfield{author}{\bibinfo{person}{Luca Zanella}, \bibinfo{person}{Willi Menapace}, \bibinfo{person}{Massimiliano Mancini}, \bibinfo{person}{Yiming Wang}, {and} \bibinfo{person}{Elisa Ricci}.} \bibinfo{year}{2024}\natexlab{}.
\newblock \showarticletitle{Harnessing large language models for training-free video anomaly detection}.
\newblock \bibinfo{journal}{{\em IEEE CVPR\/}} (\bibinfo{year}{2024}).
\newblock


\bibitem[\protect\citeauthoryear{Zhang, Ananthanarayanan, Bodik, Philipose, Bahl, and Freedman}{Zhang et~al\mbox{.}}{2017}]%
        {zhang2017videoStorm}
\bibfield{author}{\bibinfo{person}{Haoyu Zhang}, \bibinfo{person}{Ganesh Ananthanarayanan}, \bibinfo{person}{Peter Bodik}, \bibinfo{person}{Matthai Philipose}, \bibinfo{person}{Paramvir Bahl}, {and} \bibinfo{person}{Michael~J Freedman}.} \bibinfo{year}{2017}\natexlab{}.
\newblock \showarticletitle{Live video analytics at scale with approximation and $\{$Delay-Tolerance$\}$}.
\newblock \bibinfo{journal}{{\em USENIX NSDI\/}} (\bibinfo{year}{2017}).
\newblock


\bibitem[\protect\citeauthoryear{Zhang, Li, Long, Zhang, Lin, Yang, Xie, Yang, Liu, Lin, et~al\mbox{.}}{Zhang et~al\mbox{.}}{2025}]%
        {zhang2025qwen3embedding}
\bibfield{author}{\bibinfo{person}{Yanzhao Zhang}, \bibinfo{person}{Mingxin Li}, \bibinfo{person}{Dingkun Long}, \bibinfo{person}{Xin Zhang}, \bibinfo{person}{Huan Lin}, \bibinfo{person}{Baosong Yang}, \bibinfo{person}{Pengjun Xie}, \bibinfo{person}{An Yang}, \bibinfo{person}{Dayiheng Liu}, \bibinfo{person}{Junyang Lin}, {et~al\mbox{.}}} \bibinfo{year}{2025}\natexlab{}.
\newblock \showarticletitle{Qwen3 Embedding: Advancing Text Embedding and Reranking Through Foundation Models}.
\newblock \bibinfo{journal}{{\em arXiv preprint arXiv:2506.05176\/}} (\bibinfo{year}{2025}).
\newblock


\bibitem[\protect\citeauthoryear{Zhang, Zhang, Ananthanarayanan, Iyer, Shu, Bahl, Mao, and Chowdhury}{Zhang et~al\mbox{.}}{2024}]%
        {zhang2024vulcan}
\bibfield{author}{\bibinfo{person}{Yiwen Zhang}, \bibinfo{person}{Xumiao Zhang}, \bibinfo{person}{Ganesh Ananthanarayanan}, \bibinfo{person}{Anand Iyer}, \bibinfo{person}{Yuanchao Shu}, \bibinfo{person}{Victor Bahl}, \bibinfo{person}{Z~Morley Mao}, {and} \bibinfo{person}{Mosharaf Chowdhury}.} \bibinfo{year}{2024}\natexlab{}.
\newblock \showarticletitle{Vulcan: Automatic Query Planning for Live $\{$ML$\}$ Analytics}.
\newblock \bibinfo{journal}{{\em USENIX NSDI\/}} (\bibinfo{year}{2024}).
\newblock


\bibitem[\protect\citeauthoryear{Zhao, Zhang, Guo, Li, Chen, Zhao, and Huang}{Zhao et~al\mbox{.}}{2025}]%
        {zhao2025smarthome}
\bibfield{author}{\bibinfo{person}{Xinyi Zhao}, \bibinfo{person}{Congjing Zhang}, \bibinfo{person}{Pei Guo}, \bibinfo{person}{Wei Li}, \bibinfo{person}{Lin Chen}, \bibinfo{person}{Chaoyue Zhao}, {and} \bibinfo{person}{Shuai Huang}.} \bibinfo{year}{2025}\natexlab{}.
\newblock \showarticletitle{SmartHome-Bench: A Comprehensive Benchmark for Video Anomaly Detection in Smart Homes Using Multi-Modal Large Language Models}.
\newblock \bibinfo{journal}{{\em IEEE CVPR\/}} (\bibinfo{year}{2025}).
\newblock


\bibitem[\protect\citeauthoryear{Zheng, Chen, Qian, Shi, Shu, and Chen}{Zheng et~al\mbox{.}}{2024}]%
        {zheng2024review}
\bibfield{author}{\bibinfo{person}{Yue Zheng}, \bibinfo{person}{Yuhao Chen}, \bibinfo{person}{Bin Qian}, \bibinfo{person}{Xiufang Shi}, \bibinfo{person}{Yuanchao Shu}, {and} \bibinfo{person}{Jiming Chen}.} \bibinfo{year}{2024}\natexlab{}.
\newblock \showarticletitle{A review on edge large language models: Design, execution, and applications}.
\newblock \bibinfo{journal}{{\it Comput. Surveys}} (\bibinfo{year}{2024}).
\newblock


\bibitem[\protect\citeauthoryear{Zhu, Cai, Deng, Ooi, and Wu}{Zhu et~al\mbox{.}}{2024}]%
        {zhu2024llms_prompt}
\bibfield{author}{\bibinfo{person}{Jiaqi Zhu}, \bibinfo{person}{Shaofeng Cai}, \bibinfo{person}{Fang Deng}, \bibinfo{person}{Beng~Chin Ooi}, {and} \bibinfo{person}{Junran Wu}.} \bibinfo{year}{2024}\natexlab{}.
\newblock \showarticletitle{Do LLMs Understand Visual Anomalies? Uncovering LLM's Capabilities in Zero-shot Anomaly Detection}.
\newblock \bibinfo{journal}{{\em ACM MM\/}} (\bibinfo{year}{2024}).
\newblock


\bibitem[\protect\citeauthoryear{Zong, Zhang, An, Li, Xu, Xu, Tu, Xing, and Dabeer}{Zong et~al\mbox{.}}{2025}]%
        {zong2025ground}
\bibfield{author}{\bibinfo{person}{Yongshuo Zong}, \bibinfo{person}{Qin Zhang}, \bibinfo{person}{Dongsheng An}, \bibinfo{person}{Zhihua Li}, \bibinfo{person}{Xiang Xu}, \bibinfo{person}{Linghan Xu}, \bibinfo{person}{Zhuowen Tu}, \bibinfo{person}{Yifan Xing}, {and} \bibinfo{person}{Onkar Dabeer}.} \bibinfo{year}{2025}\natexlab{}.
\newblock \showarticletitle{Ground-V: Teaching VLMs to Ground Complex Instructions in Pixels}.
\newblock \bibinfo{journal}{{\em IEEE CVPR\/}} (\bibinfo{year}{2025}).
\newblock


\end{thebibliography}

\balance
\end{document}